\documentclass[10pt,twocolumn,letterpaper]{article}

\usepackage[pagenumbers]{cvpr}   %

\usepackage{cuted}
\usepackage{caption}
\usepackage{nicefrac}
\usepackage{multirow}

\usepackage{colortbl}
\definecolor{tabfirstc}{HTML}{6E9FDB}
\definecolor{tabsecondc}{HTML}{CFE1F7}
\newcommand{\tabfirst}{\cellcolor{tabfirstc}}

\definecolor{cvprblue}{rgb}{0.21,0.49,0.74}
\usepackage[pagebackref,breaklinks,colorlinks,allcolors=cvprblue]{hyperref}

\title{ATS: Anti-Aliased Triangle Splatting with Exact Per-Primitive Integrals}

\author{
Brent Zoomers\qquad
Joni Vanherck\qquad
Leander Winters\qquad
Jeroen Put\\ \quad
Lode Jorissen \qquad 
Nick Michiels\\
\small Digital Future Lab, Flanders Make, Hasselt University\qquad
}

\begin{document}
\maketitle

\begin{strip}
    \vspace{-15mm}
    \centering
    \includegraphics[width=\textwidth]{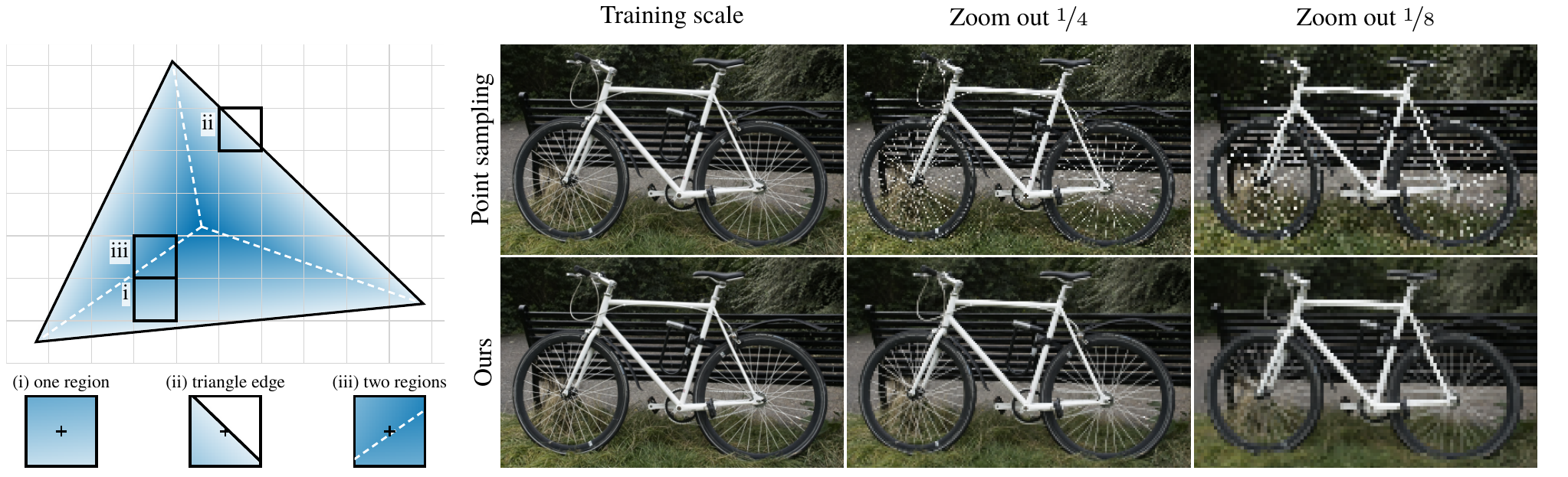}
    \captionsetup{type=figure}
    \captionof{figure}{Left: a pixel (i) inside one sub-triangle, (ii) on the triangle edge, (iii) across two sub-triangles. In each case, the integral is a sum of closed-form terms over the sides of the clipped region. Top: Triangle Splatting~\cite{held2026triangle} with point sampling and its sub-pixel cull disabled (with it, the scene disappears, \cref{fig:failure}). Bottom: ours, the exact integral of each triangle's window over the pixel. }
    \label{fig:teaser}
\end{strip}

\begin{abstract}
Triangle Splatting represents scenes using semi-transparent triangles, but it evaluates each triangle at the pixel center and culls any triangle with an inradius smaller than one pixel.
The former causes severe aliasing, while the latter culls most of the scene at lower resolutions.
We show that the integral of the Triangle Splatting window over a pixel admits an exact closed-form solution.
The angle bisectors split the triangle into three sub-triangles, in each of which the window is a power of the distance to one edge.
Each piece therefore has a closed-form antiderivative, and Green's theorem reduces the integral and its gradients to one term per side of the sub-triangle clipped against the pixel.
The integral over the whole triangle, which we show has a closed form, then allows conservative culling, removing only triangles that cannot meet the renderer's alpha cutoff at any pixel.
This exact formulation anti-aliases existing models without retraining, exhibits less temporal aliasing than $2{\times}2$ supersampling, and enables multi-scale training, matching or outperforming supersampled training at reduced resolutions.

\end{abstract}\newpage
    
\section{Introduction}
\label{sec:intro}

Triangle Splatting~\cite{held2026triangle} has been proposed as an alternative to 3D Gaussian Splatting~\cite{kerbl3Dgaussians} (3DGS), representing scenes with semi-transparent triangle primitives rather than 3D Gaussians.
They can be optimized to be fully opaque triangles, enabling hardware-accelerated rendering.
Subsequent work has already explored this primitive for downstream applications such as mesh extraction~\cite{Held_2026_CVPR} and feed-forward reconstruction~\cite{kupyn2026flatfeedforwardlatenttriangle, wang2026trisplatsimulationreadyfeedforward3d}. 
These developments highlight the potential of triangle-based representations as a versatile alternative to existing radiance field representations.

\begin{figure*}[t!]
\centering
\includegraphics[width=1.0\textwidth]{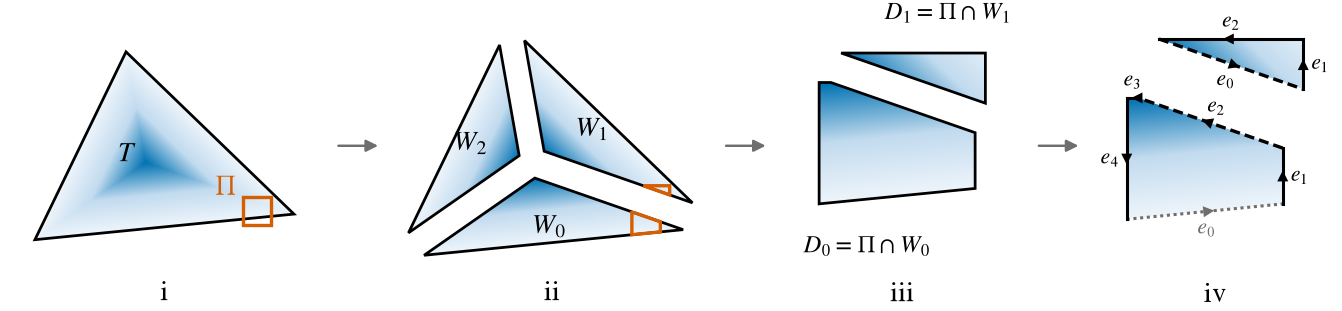}
\caption{\textbf{Pipeline per triangle.} (i) Given a triangle $T$ and a pixel $\Pi$, we aim to compute the integral of the window function inside the pixel. (ii) The bisectors split the triangle into sub-triangles $W_0, W_1, W_2$, in each of which the window depends only on the distance to one edge. (iii) Each sub-triangle is clipped to the pixel, giving $D_i = \Pi \cap W_i$. (iv) Green's theorem turns the integral over $D_i$ into one closed-form term per side $e_i$.}
\label{fig:pipeline}
\end{figure*}

Despite these advantages, triangle primitives introduce many of the same challenges as other rasterization-based rendering methods, such as 3D Gaussian Splatting~\cite{kerbl3Dgaussians}.
One such fundamental challenge is aliasing, which occurs when continuous scene information is sampled below the Nyquist rate~\cite{5055024}.
In the original Triangle Splatting~\cite{held2026triangle} formulation, primitive contributions are evaluated at pixel centers, without explicitly accounting for the full pixel footprint.
Consequently, small geometric features may be insufficiently sampled, resulting in visual artifacts that are particularly apparent when rendering resolution or viewing distance changes.
These issues limit the ability to consistently render triangle-based representations across different scales and motivate our approach to introduce anti-aliasing.
Anti-aliasing has been studied extensively in computer graphics for opaque triangles, with solutions including supersampling, prefiltering, and closed-form integration.
However, techniques developed for opaque triangles do not necessarily transfer directly to Triangle Splatting~\cite{held2026triangle}, where a primitive's transparency at each point is determined by its window function across the primitive. 
Supersampling offers one possible solution, but it requires multiple samples per pixel.
Additionally, because primitives can be smaller than the supersampling frequency, aliasing can still occur depending on the number of samples.
An analytical formulation, in contrast, offers the possibility of directly accounting for the pixel footprint while avoiding the need for dense sampling. 
The challenge is to derive such a formulation for triangle primitives while retaining the efficiency required for practical rendering.
In this work, we derive the closed-form integral of the Triangle Splatting~\cite{held2026triangle} window over a pixel, which equals the window filtered exactly with a box pixel filter.
We then use it for anti-aliasing and to replace the inradius cull with a principled culling approach. 
A key observation underlying our approach is that the Triangle Splatting window is a piecewise function where each piece has a closed-form solution: the angle bisectors split each triangle into three sub-triangles, and inside each one the window depends only on the distance to a single edge.
Green's theorem then reduces the integral to one closed-form term per side of the clipped window, and the gradients follow in the same way (see \cref{fig:pipeline}). 
This results in a closed-form solution that keeps the integral sufficiently cheap for a tile-based rasterizer at real-time frame rates.

Finally, because the per-primitive integral is exact, it is ruled out as a source of error, allowing us to evaluate remaining approximations such as the scalar blending identified in Gaussian Blending~\cite{koo2026gb}.
We quantify this error for triangles.
In summary, our contributions are:
\begin{itemize}
  \item We derive the exact closed-form integral of the Triangle Splatting window over a pixel, which results in less flickering than $2{\times}2$ supersampling, and its gradients, which improve multi-scale training.
  \item We propose a principled and conservative culling formulation that uses the lightweight integral over the full triangle to cull triangles for which no clipped part will exceed the contribution threshold used during rasterization.
  \item We integrate our method into CUDA kernels and use its accuracy to measure errors in other pipeline components.
\end{itemize}

\section{Related work}
\label{sec:related_work}

\begin{figure}[t]
  \centering
  \begin{subfigure}[t]{0.49\linewidth}
    \includegraphics[width=\linewidth]{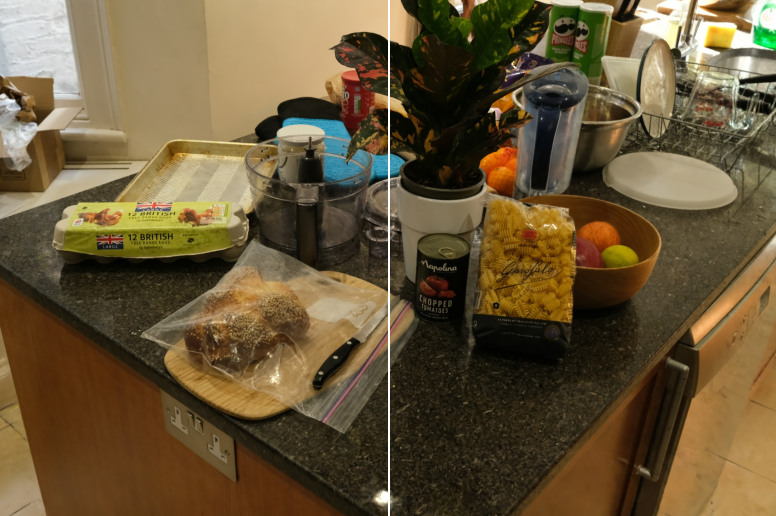}
    \caption{}
    \label{fig:cull_full}
  \end{subfigure}\hfill
  \begin{subfigure}[t]{0.49\linewidth}
    \includegraphics[width=\linewidth]{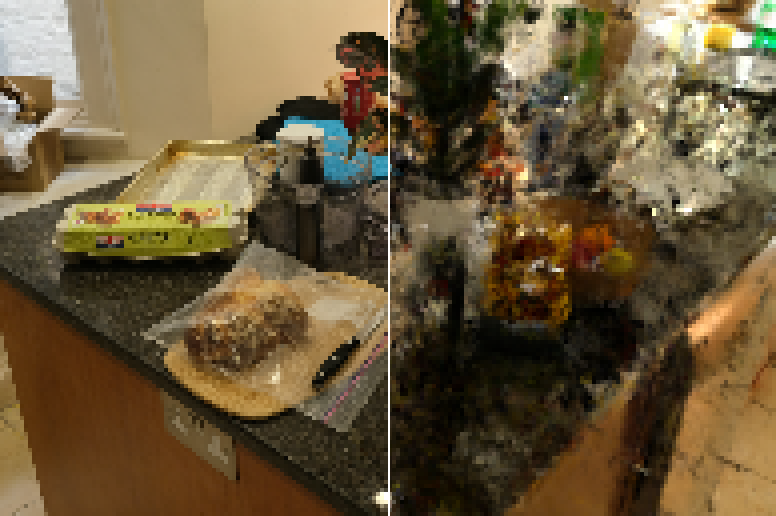}
    \caption{}
    \label{fig:cull_eighth}
  \end{subfigure}
  \caption{\textbf{Culling sub-pixel triangles.} Triangle Splatting at (a) the training resolution and (b) $\nicefrac{1}{8}$
  of it. Left half of each image: without culling and the right half with the released culling of triangles smaller than a
  pixel. At $\nicefrac{1}{8}$, many triangles are smaller than a pixel, so culling removes nearly the entire scene, which severely impacts PSNR (27.2\,dB $\rightarrow$ 18.0\,dB on this scene).}
  \label{fig:failure}
\end{figure}

Our work proposes an exact per-primitive integral over the pixel window for triangle primitives. We provide an overview of related works in novel view synthesis and anti-aliasing.

\subsection{Novel View Synthesis}

Novel View Synthesis (NVS) aims to reconstruct unseen views from a limited set of images. 
In recent years, various methods have been proposed to address this task.
NeRF~\cite{mildenhall2020nerf} and its follow-ups~\cite{mueller2022instant, 10204118, barron2021mipnerf, barron2023zipnerf, barron2022mipnerf360} render a neural field by volume rendering.
3DGS~\cite{kerbl3Dgaussians} instead rasterizes explicit 3D Gaussians, and follow-up work improves its densification~\cite{10.1145/3680528.3687694, 10.1007/978-3-031-73036-8_20, 11697845}, rendering speed~\cite{HansonSpeedy, Hahlbohm_2026_CVPR, pajak2026higshierarchicalrenderingarchitecture} and addresses anti-aliasing ~\cite{Yu2024MipSplatting, 10.1007/978-3-031-72643-9_17, koo2026gb, yan2024multi}.
Recent works introduce other scene representations such as convex shapes~\cite{Held20243DConvex}, constant-density ellipsoids~\cite{11444285}, and Voronoi-based methods~\cite{mai2026radiance, Govindarajan_2025_ICCV}.
One such method uses semi-transparent triangles to optimize scenes~\cite{held2026triangle}, arguing for a comeback of triangles.
Again, various follow-up approaches were proposed, ranging from improved mesh extraction~\cite{Held_2026_CVPR} to feed-forward Triangle Splatting~\cite{kupyn2026flatfeedforwardlatenttriangle, wang2026trisplatsimulationreadyfeedforward3d}.
With this primitive gaining more attention, we focus on addressing anti-aliasing, which, for this particular primitive, has an elegant solution: a closed-form, per-primitive analytical integral over the pixel area.

\subsection{Anti-aliasing}

Aliasing has been explored as a problem in the community for decades. 
Earlier works focused on addressing this issue for triangles~\cite{10.1145/359863.359869, 10.1145/800248.807360, aa_comparison, duff1989polygon}, for linear functions on polytopes~\cite{https://doi.org/10.1111/j.1467-8659.2012.03012.x}, or for curves~\cite{Manson:2013:ARC}.
The latter uses the divergence theorem, with Green's Theorem as the 2D case, while the rest focuses on triangles or, specifically, linear functions.
Our proposed method targets the window function used in Triangle Splatting~\cite{held2026triangle}, which is a power function rather than a linear or constant one.
In EWA Splatting~\cite{1021576}, anti-aliasing for elliptical Gaussian kernels is explored.
They prefilter the Gaussian after projection with a low-pass filter, which limits the representation of high-frequency components.
Since Triangle Splatting uses a drastically different window function, these methods do not carry over directly.
For NeRF~\cite{mildenhall2020nerf}, Mip-NeRF~\cite{barron2021mipnerf} addressed anti-aliasing by considering conical frustums instead of individual rays and using integrated positional encodings to allow encoding space rather than individual points.
To address anti-aliasing in 3D Gaussian Splatting~\cite{kerbl3Dgaussians}, several methods have been proposed.
First, Mip-Splatting~\cite{Yu2024MipSplatting} introduces both a 2D and a 3D filter.
While the 2D filter ensures that Gaussians do not become smaller than a pixel after projection, the 3D filter aims to limit the maximum frequency of Gaussians to a value determined by the sampling rate of the ground truth images.
Multi-scale 3DGS~\cite{yan2024multi} uses different sets of Gaussians and swaps between them based on pixel coverage to avoid sub-pixel Gaussians.
Analytic-Splatting~\cite{10.1007/978-3-031-72643-9_17} proposes an analytical approximation of the window integral, which has no closed form for a rotated anisotropic Gaussian. 
Compared to Mip-Splatting~\cite{Yu2024MipSplatting}, it does not suppress high-frequency Gaussians.
AAA-Gaussians~\cite{steiner2025aaags} propose a 3D replacement of the 2D filter in Mip-Splatting to avoid the approximation due to projection.
More recently, Gaussian Blending~\cite{koo2026gb} focused on the issue of treating opacity as a scalar value over a pixel.
A per-primitive integral over the pixel window alone does not guarantee accurate rendering, since each primitive is blended front-to-back with a per-pixel scalar value.
This does not account for opacity distributions or how primitives are distributed across the pixel.
Compared to prior work, our method proposes an exact per-primitive integral of the window function in Triangle Splatting~\cite{held2026triangle} over the pixel window and applies Green's Theorem to a power function that has a closed-form solution.

\begin{figure*}[t!]
\centering
\includegraphics[width=1.0\textwidth]{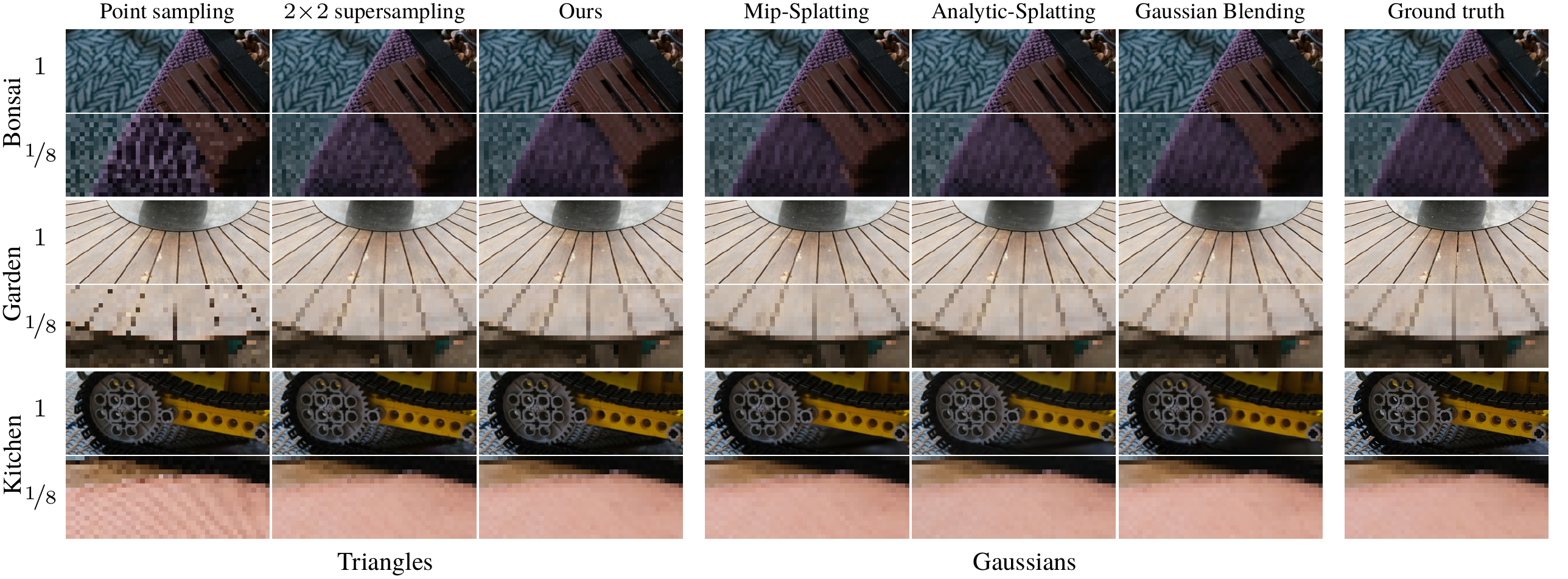}
\caption{\textbf{Zooming out after single-scale training} on three Mip-NeRF~360 scenes. Each crop shows the training resolution (top) and its $\nicefrac18$ (bottom), upsampled. The left shows triangle-based methods, the middle shows Gaussian anti-aliasing methods, and the right shows ground truth. Point sampling aliases at $\nicefrac18$; $2{\times}2$ supersampling reduces this but still leaves some aliasing, e.g., on bonsai's tablecloth. }
\label{fig:visual_comparison}
\end{figure*}

\section{Method}
\label{sec:method}

We first discuss the window function as used in Triangle Splatting~\cite{held2026triangle} in \cref{sec:prelims}. 
We then provide an overview of our method starting with the derivation of the integral in \cref{sec:exact_integral}, the updated culling formulation in \cref{sec:culling}, the custom backward pass in \cref{sec:backward}, and optimizations for efficient rendering in \cref{sec:efficient}.

\subsection{Preliminaries}
\label{sec:prelims}

\textbf{Triangle Splatting} uses projected vertices and SDFs in 2D to compute the contribution of each primitive. The SDF is based on the outward-pointing unit normals $\mathbf{n_k}$ of the projected 2D triangle, the line's offset $O_k$, and the point $\mathbf{p}$ itself, with $k$ representing the edges.

\begin{equation}
    \phi(\mathbf{p}) = \max\limits_{k}(d_k(\mathbf{p})), \quad d_k(\mathbf{p}) = \mathbf{n_k} \cdot \mathbf{p} + O_k
\end{equation}

\begin{equation}
    I(\mathbf{p}) = \text{ReLU}(\frac{-\phi(\mathbf{p})}{r})^\sigma
\end{equation}
where $r$ is the inradius and $\sigma$ is a learnable smoothness value that determines the falloff from the incenter towards the edges.
This window function ensures that the maximum value within the triangle is 1, the inside is positive, and the boundaries and outside are 0.
Notice how the maximum function divides the triangle into three sub-triangles for which it only depends on the nearest edge (see \cref{fig:pipeline}).
Furthermore, they cull every triangle whose projected inradius is below one pixel.
When the resolution is lowered, nearly the entire scene is removed, as shown in \cref{fig:failure}.

\subsection{Exact Integral for Triangle Primitive}
\label{sec:exact_integral}

Our goal is to find the exact contribution of a primitive to a pixel.
To achieve this, we integrate the window function over the pixel area by splitting the triangle into three pieces along the bisectors, where each bisector is defined as: 
\begin{equation}
    d_i(\mathbf{p}) - d_j(\mathbf{p}) = 0 
\end{equation}

Since the window function is piecewise, with the edge used to determine the value being the one closest to point $\mathbf{p}$, we integrate each sub-triangle separately and sum the results to obtain the final result.
Without loss of generality, we can thus write the function inside the power for each sub-triangle formed by the bisectors of the triangle primitive.
\begin{equation}
\label{eq:eq_u}
    u(\mathbf{p}) = \frac{-(\mathbf{n} \cdot \mathbf{p} + O)}{r} = \frac{-d(\mathbf{p})}{r}
\end{equation}
where $n$ represents the normal of the edge, and $O$ is the offset.
The window function becomes:
\begin{equation}
    I(\mathbf{p}) = u(\mathbf{p})^\sigma
\end{equation}
with $\sigma$ the smoothness value as in Triangle Splatting~\cite{held2026triangle}.
Note that we can remove the ReLU activation function, since we clip the pixel at the edge and at the two bisectors.
This means the integral we want to compute for a sub-triangle becomes the integral of the window function over the clipped pixel area $D$.
\begin{equation}
\label{eq:double_integral_dxdy}
    C = \iint_D I(\mathbf{p}) dA = \iint_D I(\mathbf{p}) dxdy
\end{equation}

To simplify the derivation, we change the variables from
$x$ and $y$ to $u$ and $a$, where $u$ is given by
\cref{eq:eq_u} and $a$ denotes the position along the edge of
the sub-triangle.
In these coordinates, the window depends only on $u$, so a single antiderivative in $u$ suffices for every edge. 
In $x$ and $y$, the antiderivative would also divide by a component of $\mathbf{n}$, which vanishes for edges parallel to an axis, so the axis of integration would have to be chosen per edge.
The functions to transform $x$ and $y$ to $u$ is \cref{eq:eq_u} and to $a$ is as follows:
\begin{equation}
\label{eq:a_transform}
  a = \mathbf{t} \cdot \mathbf{p}, \quad
  \mathbf{t} = (-n_y, n_x)
\end{equation}
where $\mathbf{t}$ denotes the normal rotated by 90 degrees counterclockwise to be aligned with the edge.
The change of variables theorem then states that $dx\,dy = r\,du\,da$, since the determinant of the Jacobian equals $(n_x^2 + n_y^2)/r = 1/r$ for a unit normal $\mathbf{n}$. Substituting into \cref{eq:double_integral_dxdy}:

\begin{equation}\label{eq:dadu_surface}
    C = r\iint_{D'} I(\mathbf{p}) dadu
\end{equation}
with $D'$ the transformed clipped area. Next, we apply Green's Theorem to convert the area integral into a contour integral.

\begin{equation}
\label{eq:green}
    \oint_E Pda + Qdu = \iint_{D'} \frac{\partial Q}{\partial a} - \frac{\partial P}{\partial u} dadu
\end{equation}

We choose $P$ and $Q$ to simplify further derivation.
Set $Q = 0$, so that $-\frac{\partial P}{\partial u} = u^\sigma$ must hold: 

\begin{equation}
    C = -\frac{r}{\sigma +1} \oint_E u(\mathbf{p})^{\sigma + 1} da
\end{equation}
Our contour consists of edges $E$ of the polygon obtained by clipping the window function to the pixel boundaries.
This contour integral thus reduces to the sum of the line integrals over the edges of the polygon.
\begin{equation}\label{eq:sumE}
    C = -\frac{r}{\sigma +1} \sum_E \int_{a_A}^{a_B} u(\mathbf{p})^{\sigma + 1} da
\end{equation}
where $E$ is traversed counter-clockwise.
Given that we have a formula in terms of $u$ and try to integrate to $a$, we rewrite $da$ in terms of $du$.
Since $a$ and $u$ both are linear in $(x,y)$ we can parameterize both with $\lambda \in [0,1]$ and solve for $da$:
\begin{equation}\label{eq:parameterization}
    \begin{gathered}
        a = a_A + \lambda (a_B - a_A), \quad
        u = u_A + \lambda (u_B - u_A) \\
        da = \frac{a_B - a_A}{u_B - u_A}\, du
    \end{gathered}
\end{equation}
We now solve the integral and simplify to achieve:

\begin{equation}\label{eq:almost_final}
    \begin{split}
    C = -\frac{r}{\sigma + 1} \sum_E (a_B - a_A) \frac{u_{B}^{\sigma + 2} - u_A^{\sigma + 2}}{(\sigma+2)(u_B - u_A)}
    \end{split}
\end{equation}

For sides with $u_A \approx u_B$, we evaluate the quotient with a Taylor expansion (supplementary material).
We then alpha-blend with $\alpha = oC$ for that triangle within the pixel, and the rest of the pipeline remains unchanged. 
As $\sigma \to 0$, the window tends to $1$ inside the triangle,
and $C$ becomes the fraction of the pixel that the triangle
covers.

\subsection{Triangle Culling}
\label{sec:culling}
We cull a triangle when the integral of its window over the whole triangle, times its opacity, falls below the alpha cutoff $\lambda_\text{cull}$.
Since no pixel integral can exceed this total, such a triangle stays below the cutoff in every pixel, and culling it has no visual impact (see supplementary material).
The total is \cref{eq:almost_final} for a pixel that contains the whole triangle, so that each clipped area $D$ is a full sub-triangle $W$.
Its boundary is the edge from vertex $H$ to vertex $K$, and the two bisector sides from $K$ to the incenter $J$ and from $J$ back to $H$.
Filling this into \cref{eq:almost_final}, $u$ remains zero on the edge, so this side adds nothing, while on the bisector sides $u$ goes from 0 to 1 or reverse, so they add $(a_J - a_K)/(\sigma + 2)$ and $(a_H - a_J)/(\sigma + 2)$.
The incenter's $a_J$ cancels in their sum, and with $a_H - a_K = -L_k$, the length of the edge, this gives:
\begin{equation}
\label{eq:sub_triangle_integral}
    C_W = -\frac{r}{\sigma + 1} \frac{a_H - a_K}{\sigma + 2} = \frac{r L_k}{(\sigma + 1)(\sigma + 2)}
\end{equation}

As $r L_k$ is twice the area of the sub-triangle, summing over the three sub-triangles gives twice the area $A$ of the full triangle in pixels, and we cull when
\begin{equation}
\label{eq:cull}
    o \cdot \frac{2A}{(\sigma + 1)(\sigma + 2)} < \lambda_\text{cull}
\end{equation}

\subsection{Backward Pass}
\label{sec:backward}

\begin{table*}[t]
\centering\small\setlength{\tabcolsep}{3pt}
\caption{\textbf{Zoom-out after single-scale training.} Test views are rendered at $1$ to $\nicefrac{1}{8}$ of the training resolution and compared with the box-downsampled ground truth, and we report the mean over the scenes of each data set. Triangle Splatting$_{\text{no culling}}$ is the released method without culling of sub-pixel triangles in training and rendering, as the cull breaks down when zooming out (supplementary material), and Triangle Splatting$_{\text{supersampled}}$ renders that model with $2{\times}2$ samples per pixel. All models are trained in our framework using the released settings and a numerically stable opacity update for cloned triangles (see supplementary material). \colorbox{tabfirstc}{Blue} marks the best value.}
\label{tab:comparison_ss}
\resizebox{\textwidth}{!}{%
\begin{tabular}{lcccccccccccc}
\toprule
Method & \multicolumn{3}{c}{$1$} & \multicolumn{3}{c}{$\nicefrac{1}{2}$} & \multicolumn{3}{c}{$\nicefrac{1}{4}$} & \multicolumn{3}{c}{$\nicefrac{1}{8}$} \\
\cmidrule(lr){2-4}\cmidrule(lr){5-7}\cmidrule(lr){8-10}\cmidrule(lr){11-13}
 & PSNR$^{\uparrow}$ & SSIM$^{\uparrow}$ & LPIPS$^{\downarrow}$ & PSNR$^{\uparrow}$ & SSIM$^{\uparrow}$ & LPIPS$^{\downarrow}$ & PSNR$^{\uparrow}$ & SSIM$^{\uparrow}$ & LPIPS$^{\downarrow}$ & PSNR$^{\uparrow}$ & SSIM$^{\uparrow}$ & LPIPS$^{\downarrow}$ \\
\midrule
\multicolumn{13}{l}{\textbf{\emph{Mip-NeRF 360}}} \\
Triangle Splatting$_{\text{no culling}}$ & 27.16 & 0.815 & 0.190 & 27.78 & 0.850 & 0.121 & 26.87 & 0.827 & 0.130 & 24.83 & 0.765 & 0.178 \\
Triangle Splatting$_{\text{supersampled}}$ & \tabfirst{27.27} & \tabfirst{0.819} & 0.192 & \tabfirst{28.41} & \tabfirst{0.868} & \tabfirst{0.113} & \tabfirst{29.69} & \tabfirst{0.904} & \tabfirst{0.072} & \tabfirst{29.90} & 0.908 & 0.072 \\
Ours & 27.26 & \tabfirst{0.819} & \tabfirst{0.187} & 28.36 & 0.866 & \tabfirst{0.113} & 29.33 & 0.902 & 0.075 & 29.26 & \tabfirst{0.920} & \tabfirst{0.062} \\
\midrule
\multicolumn{13}{l}{\textbf{\emph{Tanks and Temples}}} \\
Triangle Splatting$_{\text{no culling}}$ & 23.14 & 0.860 & 0.140 & 23.54 & 0.888 & 0.092 & 22.57 & 0.864 & 0.115 & 20.83 & 0.823 & 0.156 \\
Triangle Splatting$_{\text{supersampled}}$ & 23.24 & 0.863 & 0.142 & \tabfirst{24.02} & \tabfirst{0.903} & \tabfirst{0.081} & \tabfirst{24.78} & \tabfirst{0.928} & \tabfirst{0.057} & \tabfirst{24.58} & 0.926 & 0.066 \\
Ours & \tabfirst{23.30} & \tabfirst{0.864} & \tabfirst{0.136} & \tabfirst{24.02} & \tabfirst{0.903} & \tabfirst{0.081} & 24.56 & 0.925 & 0.059 & 24.45 & \tabfirst{0.931} & \tabfirst{0.056} \\
\midrule
\multicolumn{13}{l}{\textbf{\emph{Deep Blending}}} \\
Triangle Splatting$_{\text{no culling}}$ & 28.94 & 0.893 & 0.246 & 29.19 & 0.897 & 0.167 & 29.27 & 0.901 & 0.128 & 28.28 & 0.891 & 0.136 \\
Triangle Splatting$_{\text{supersampled}}$ & \tabfirst{28.98} & \tabfirst{0.895} & 0.246 & \tabfirst{29.32} & \tabfirst{0.901} & 0.167 & \tabfirst{29.94} & \tabfirst{0.915} & \tabfirst{0.109} & \tabfirst{30.56} & \tabfirst{0.928} & \tabfirst{0.086} \\
Ours & 28.91 & 0.894 & \tabfirst{0.244} & 29.26 & 0.900 & \tabfirst{0.166} & 29.77 & 0.914 & 0.110 & 30.04 & 0.926 & 0.087 \\
\midrule
\multicolumn{13}{l}{\textbf{\emph{NeRF Synthetic}}} \\
Triangle Splatting$_{\text{no culling}}$ & 33.39 & \tabfirst{0.969} & \tabfirst{0.031} & 33.40 & 0.973 & 0.023 & 29.97 & 0.964 & 0.040 & 26.34 & 0.944 & 0.074 \\
Triangle Splatting$_{\text{supersampled}}$ & 33.40 & \tabfirst{0.969} & 0.032 & \tabfirst{35.38} & \tabfirst{0.978} & \tabfirst{0.018} & \tabfirst{35.78} & \tabfirst{0.982} & \tabfirst{0.015} & \tabfirst{33.17} & \tabfirst{0.980} & \tabfirst{0.021} \\
Ours & \tabfirst{33.55} & \tabfirst{0.969} & \tabfirst{0.031} & 34.64 & 0.977 & \tabfirst{0.018} & 33.29 & 0.980 & 0.018 & 30.40 & 0.977 & 0.023 \\
\bottomrule
\end{tabular}}
\end{table*}

Compared to Triangle Splatting, only $\alpha = o\,C$ changes, so we need the derivatives of $C$ with respect to the input parameters.
We write $C = \sum_W C_W$, with $C_W$ the integral over the part $D$ of the pixel inside sub-triangle $W$.
Applying the Leibniz integral rule with variable limits moves the derivative inside the integral and adds the integrand, evaluated at the moving limits, multiplied by their rate of change.
We show in the supplementary material that these extra terms are zero or cancel for the Triangle Splatting window, leaving only the integral of the derivative of the integrand.
The derivatives with respect to the $n$ and $O$ of an edge therefore come only from its own sub-triangle, and $\partial C / \partial \sigma = \sum_W \partial C_W / \partial \sigma$.
\begin{table}[b]
\centering\small\setlength{\tabcolsep}{3pt}
\caption{\textbf{Temporal stability.} Error of the frame-to-frame differences ($\times 10^{-3}$, lower is better) against a $16{\times}16$ supersampled render of the same model, at $\nicefrac{1}{4}$ and $\nicefrac{1}{8}$ of the training resolution. All renderers draw the Triangle Splatting$_{\text{no culling}}$ model of \cref{tab:comparison_ss}. We report the mean over all scenes. \colorbox{tabfirstc}{Blue} marks the best value.}
\label{tab:flicker_metric}
\resizebox{\columnwidth}{!}{%
\begin{tabular}{lcccccc}
\toprule
 & \multicolumn{2}{c}{Mip-NeRF 360} & \multicolumn{2}{c}{Tanks and Temples} & \multicolumn{2}{c}{Deep Blending} \\
\cmidrule(lr){2-3}\cmidrule(lr){4-5}\cmidrule(lr){6-7}
Renderer & $\nicefrac{1}{4}$ & $\nicefrac{1}{8}$ & $\nicefrac{1}{4}$ & $\nicefrac{1}{8}$ & $\nicefrac{1}{4}$ & $\nicefrac{1}{8}$ \\
Point sampling & 10.1 & 20.7 & 13.1 & 27.0 & 3.7 & 8.6 \\
$2{\times}2$ supersampling & 3.8 & 9.2 & 5.1 & 13.0 & 1.4 & 3.7 \\
\midrule
Ours & \tabfirst{1.9} & \tabfirst{3.3} & \tabfirst{2.7} & \tabfirst{4.5} & \tabfirst{0.9} & \tabfirst{1.7} \\
\bottomrule
\end{tabular}}
\end{table}

With the inradius $r$ treated as a constant as in Triangle Splatting (supplementary material),
\begin{equation}
\begin{aligned}
  \frac{\partial C_W}{\partial O} &= -\frac{\sigma}{r} \iint_D u^{\sigma-1}\, dx\, dy, \\
  \frac{\partial C_W}{\partial \mathbf{n}} &= -\frac{\sigma}{r} \iint_D \mathbf{p}\, u^{\sigma-1}\, dx\, dy, \\
  \frac{\partial C_W}{\partial \sigma} &= \iint_D u^{\sigma} \ln u\, dx\, dy .
\end{aligned}
\label{eq:grad_area}
\end{equation}
Each is of the same kind as $C_W$, so Green's theorem reduces it to one closed-form term per side $E$ of $D$, as in the forward pass, and one pass over the sides gives both $C$ and its gradients.
The gradients with respect to $n$ and $O$ are propagated to the projected vertices by the chain rule, as in Triangle Splatting~\cite{held2026triangle}.

\begin{table*}[t]
\centering\small\setlength{\tabcolsep}{3pt}
\caption{\textbf{Zoom-out after multi-scale training} with the scenes of \cref{tab:comparison_ss}, and on NeRF Synthetic at the four scales of its multi-scale version~\cite{barron2021mipnerf}. Each iteration renders a training view at a random scale from $1$ to $\nicefrac{1}{8}$, with full resolution forced in $30\%$ of the iterations as in Mip-Splatting~\cite{Yu2024MipSplatting}, and every model is trained with the renderer it is tested with. Triangle Splatting$_{\text{no culling}}$ has no NeRF Synthetic row, since its training ran out of memory on lego, and Triangle Splatting$_{\text{supersampled}}$ is trained and rendered with $2{\times}2$ samples per pixel. \colorbox{tabfirstc}{Blue} marks the best value.}
\label{tab:comparison_ms}
\resizebox{\textwidth}{!}{%
\begin{tabular}{lcccccccccccc}
\toprule
Method & \multicolumn{3}{c}{$1$} & \multicolumn{3}{c}{$\nicefrac{1}{2}$} & \multicolumn{3}{c}{$\nicefrac{1}{4}$} & \multicolumn{3}{c}{$\nicefrac{1}{8}$} \\
\cmidrule(lr){2-4}\cmidrule(lr){5-7}\cmidrule(lr){8-10}\cmidrule(lr){11-13}
 & PSNR$^{\uparrow}$ & SSIM$^{\uparrow}$ & LPIPS$^{\downarrow}$ & PSNR$^{\uparrow}$ & SSIM$^{\uparrow}$ & LPIPS$^{\downarrow}$ & PSNR$^{\uparrow}$ & SSIM$^{\uparrow}$ & LPIPS$^{\downarrow}$ & PSNR$^{\uparrow}$ & SSIM$^{\uparrow}$ & LPIPS$^{\downarrow}$ \\
\midrule
\multicolumn{13}{l}{\emph{Mip-NeRF 360}} \\
Triangle Splatting$_{\text{no culling}}$ & 25.69 & 0.735 & 0.293 & 26.87 & 0.815 & 0.170 & 27.82 & 0.864 & 0.106 & 26.45 & 0.835 & 0.130 \\
Triangle Splatting$_{\text{supersampled}}$ & 26.65 & 0.784 & 0.236 & 27.98 & 0.854 & 0.133 & 29.56 & 0.905 & 0.074 & 30.27 & 0.921 & 0.063 \\
Ours & \tabfirst{26.91} & \tabfirst{0.802} & \tabfirst{0.218} & \tabfirst{28.29} & \tabfirst{0.865} & \tabfirst{0.121} & \tabfirst{29.93} & \tabfirst{0.913} & \tabfirst{0.068} & \tabfirst{31.00} & \tabfirst{0.941} & \tabfirst{0.046} \\
\midrule
\multicolumn{13}{l}{\emph{Tanks and Temples}} \\
Triangle Splatting$_{\text{no culling}}$ & 22.21 & 0.802 & 0.238 & 23.05 & 0.864 & 0.137 & 23.22 & 0.885 & 0.105 & 21.90 & 0.861 & 0.126 \\
Triangle Splatting$_{\text{supersampled}}$ & 22.85 & 0.835 & 0.188 & 23.77 & 0.889 & 0.102 & 24.68 & 0.923 & 0.064 & 24.77 & 0.931 & 0.062 \\
Ours & \tabfirst{22.99} & \tabfirst{0.847} & \tabfirst{0.168} & \tabfirst{23.91} & \tabfirst{0.896} & \tabfirst{0.093} & \tabfirst{24.85} & \tabfirst{0.929} & \tabfirst{0.058} & \tabfirst{25.43} & \tabfirst{0.946} & \tabfirst{0.044} \\
\midrule
\multicolumn{13}{l}{\emph{Deep Blending}} \\
Triangle Splatting$_{\text{no culling}}$ & 28.08 & 0.879 & 0.295 & 28.40 & 0.888 & 0.206 & 28.85 & 0.902 & 0.136 & 28.40 & 0.902 & 0.124 \\
Triangle Splatting$_{\text{supersampled}}$ & \tabfirst{28.62} & \tabfirst{0.889} & 0.265 & \tabfirst{29.00} & 0.899 & 0.179 & 29.66 & 0.915 & 0.110 & \tabfirst{30.36} & 0.929 & 0.084 \\
Ours & 28.57 & \tabfirst{0.889} & \tabfirst{0.264} & 28.99 & \tabfirst{0.900} & \tabfirst{0.175} & \tabfirst{29.68} & \tabfirst{0.917} & \tabfirst{0.108} & \tabfirst{30.36} & \tabfirst{0.931} & \tabfirst{0.080} \\
\midrule
\multicolumn{13}{l}{\emph{NeRF Synthetic}} \\
Triangle Splatting$_{\text{supersampled}}$ & 30.39 & 0.953 & 0.055 & 32.42 & 0.968 & 0.029 & 34.23 & 0.976 & 0.020 & 33.08 & 0.978 & 0.021 \\
Ours & \tabfirst{31.52} & \tabfirst{0.960} & \tabfirst{0.046} & \tabfirst{33.61} & \tabfirst{0.974} & \tabfirst{0.023} & \tabfirst{35.41} & \tabfirst{0.982} & \tabfirst{0.014} & \tabfirst{34.98} & \tabfirst{0.986} & \tabfirst{0.012} \\
\bottomrule
\end{tabular}}
\end{table*}

\subsection{Efficient rendering}
\label{sec:efficient}
The pixel integral costs more per triangle and pixel than a point sample, and at low resolution a frame has too few $16{\times}16$ tiles to occupy the GPU (70--117 at $\nicefrac18$ on Mip-NeRF~360, for 170 multiprocessors).
We therefore render $8{\times}8$ tiles and cut each tile's depth-sorted triangle list into slices of similar length, each rendered by its own thread block. The slices' colors and transmittances are then composited front-to-back, which changes the PSNR by at most $0.001$\,dB.
In training, a first pass computes the transmittance at the start of each slice, so that the second pass and the backward pass start each slice from the right state.
When a pixel lies entirely within a single sub-triangle, we integrate only over the four sides of the pixel.

\section{Evaluation}
\label{sec:evaluation}

We first compare our method against Triangle Splatting and supersampling to show its performance under single- and multi-scale training in \cref{sec:comparison}.
Next, we compare against Gaussian-based methods. After single-scale training, our PSNR at $\nicefrac{1}{8}$ is higher than that of the established anti-aliasing methods Mip-Splatting~\cite{Yu2024MipSplatting} and Analytic-Splatting~\cite{10.1007/978-3-031-72643-9_17}.
Gaussian Blending~\cite{koo2026gb}, which addresses per-pixel compositing, achieves higher metrics in this setting, so we isolate the impact of compositing in \cref{sec:analysis}.

\begin{table}[b]
\centering\small\setlength{\tabcolsep}{3pt}
\caption{\textbf{Inference speed} in frames per second (FPS) of the single-scale
models of \cref{tab:comparison_ss}, mean over the nine Mip-NeRF~360 scenes
(RTX 5090, median over test views). $2{\times}2$ supersampling
draws the Triangle Splatting$_{\text{no culling}}$ model. }
\label{tab:speed}
\begin{tabular*}{\linewidth}{@{\extracolsep{\fill}}lcccc@{}}
\toprule
Renderer & $1$ & $\nicefrac{1}{2}$ & $\nicefrac{1}{4}$ & $\nicefrac{1}{8}$ \\
\midrule
Triangle Splatting$_{\text{no culling}}$ & 202 & 218 & 143 & 69 \\
\quad + depth split, $8{\times}8$ tiles & 197 & 287 & 357 & 397 \\
$2{\times}2$ supersampling & 108 & 204 & 220 & 144 \\
\quad + depth split, $8{\times}8$ tiles  & 101 & 187 & 280 & 350 \\
Ours, exact & 81 & 97 & 56 & 27 \\
\quad + depth split, $8{\times}8$ tiles & 99 & 164 & 238 & 322 \\
\bottomrule
\end{tabular*}
\end{table}

\subsection{Datasets \& Metrics}
\label{sec:eval_protocol} 

We use all nine Mip-NeRF 360~\cite{barron2022mipnerf360} scenes, with the outdoor scenes at $\nicefrac{1}{4}$ and the indoor scenes at $\nicefrac{1}{2}$ of the original resolution, \emph{truck} and \emph{train} from Tanks and Temples~\cite{Knapitsch2017}, \emph{drjohnson} and \emph{playroom} from Deep Blending~\cite{deep_blending}, as in 3DGS~\cite{kerbl3Dgaussians}, and the eight NeRF Synthetic~\cite{mildenhall2020nerf} scenes.
Test views are rendered at $1$ to $\nicefrac{1}{8}$ of the training resolution and compared with the box-downsampled ground truth. 
The Gaussian methods are trained with the authors' code and settings, and all methods are evaluated with the same protocol. 
We use standard image metrics PSNR, LPIPS~\cite{8578166}, and SSIM~\cite{article_SSIM} to assess image quality, and we measure FPS for all methods. 
We measure flickering as the error between frame-to-frame differences and a $16{\times}16$ supersampled render of the same model, over 128 frames, each shifted by $\nicefrac{1}{16}$ of a pixel, across three test views per scene.

\begin{table*}[t]
\centering\small\setlength{\tabcolsep}{3pt}
\caption{\textbf{Comparison with Gaussian-based methods} after single-scale training, at the training resolution ($1$) and at $\nicefrac{1}{8}$ of it, with the scenes and resolutions of \cref{tab:comparison_ss}. The Gaussian-based methods are trained with the authors' code and settings and evaluated with the same protocol as ours. All scales, data sets and multi-scale training are in the supplementary material. \colorbox{tabfirstc}{Blue} marks the best value.}
\label{tab:comparison_gaussians}
\resizebox{\textwidth}{!}{%
\begin{tabular}{lcccccccccccc}
\toprule
 & \multicolumn{6}{c}{Mip-NeRF 360} & \multicolumn{6}{c}{Tanks and Temples} \\
\cmidrule(lr){2-7}\cmidrule(lr){8-13}
Method & \multicolumn{3}{c}{$1$} & \multicolumn{3}{c}{$\nicefrac{1}{8}$} & \multicolumn{3}{c}{$1$} & \multicolumn{3}{c}{$\nicefrac{1}{8}$} \\
\cmidrule(lr){2-4}\cmidrule(lr){5-7}\cmidrule(lr){8-10}\cmidrule(lr){11-13}
 & PSNR$^{\uparrow}$ & SSIM$^{\uparrow}$ & LPIPS$^{\downarrow}$ & PSNR$^{\uparrow}$ & SSIM$^{\uparrow}$ & LPIPS$^{\downarrow}$ & PSNR$^{\uparrow}$ & SSIM$^{\uparrow}$ & LPIPS$^{\downarrow}$ & PSNR$^{\uparrow}$ & SSIM$^{\uparrow}$ & LPIPS$^{\downarrow}$ \\
\midrule
3DGS~\cite{kerbl3Dgaussians} & 27.52 & 0.814 & 0.217 & 18.79 & 0.567 & 0.240 & \tabfirst{23.87} & 0.851 & 0.170 & 17.10 & 0.619 & 0.207 \\
Mip-Splatting~\cite{Yu2024MipSplatting} & \tabfirst{27.66} & \tabfirst{0.827} & 0.190 & 28.89 & 0.925 & \tabfirst{0.060} & 23.79 & 0.858 & 0.157 & 24.07 & 0.919 & 0.062 \\
Analytic-Splatting~\cite{10.1007/978-3-031-72643-9_17} & 27.52 & 0.816 & 0.215 & 28.47 & 0.917 & 0.074 & 23.75 & 0.853 & 0.173 & 23.78 & 0.912 & 0.067 \\
Gaussian Blending~\cite{koo2026gb} & 26.86 & 0.803 & 0.237 & \tabfirst{30.37} & \tabfirst{0.931} & 0.068 & 23.27 & 0.846 & 0.188 & \tabfirst{25.41} & \tabfirst{0.936} & \tabfirst{0.053} \\
Ours & 27.26 & 0.819 & \tabfirst{0.187} & 29.26 & 0.920 & 0.062 & 23.30 & \tabfirst{0.864} & \tabfirst{0.136} & 24.45 & 0.931 & 0.056 \\
\bottomrule
\end{tabular}}
\end{table*}

\subsection{Implementation details}

We train for 30{,}000 iterations with Triangle Splatting's~\cite{held2026triangle} hyperparameters, except that we replace their inradius cull with ours using $\lambda_\text{cull} = \nicefrac1{255}$, the alpha cutoff of the 3DGS and Triangle Splatting rasterizers.
All triangle models that we train use a numerically stable form of Triangle Splatting's opacity update for cloned triangles.
For Mip-Splatting~\cite{Yu2024MipSplatting}, we use their latest public code release, which achieves higher image metrics due to using improved densification of Gaussian Opacity Fields~\cite{Yu2024GOF}.
All models were trained on an NVIDIA RTX 5090 using standard training/test splits.

\subsection{Comparative Evaluation}
\label{sec:comparison}

\Cref{tab:comparison_ss} trains every model at the resolution of \cref{sec:eval_protocol} and renders the test views at $1$ to $\nicefrac{1}{8}$ of it.
At $\nicefrac{1}{8}$ on Mip-NeRF~360, ours reaches $29.26$\,dB, against $24.83$\,dB for Triangle Splatting$_{\text{no culling}}$, and it is on par with both at the training resolution.
Both supersampling and ours outperform Triangle Splatting$_{\text{no culling}}$  at all lower resolutions. 
Supersampling beats our PSNR in most cases (\cref{sec:analysis}), while at $\nicefrac{1}{8}$ ours achieves better SSIM and LPIPS on Mip-NeRF~360 and Tanks and Temples.
Additionally, our method achieves significantly less flickering.
On a pan that moves the image by $\nicefrac{1}{16}$ pixel per frame, the error of frame-to-frame differences at $\nicefrac{1}{8}$ on Mip-NeRF~360 drops from $9.2$ with $2{\times}2$ supersampling to $3.3$ with ours (\cref{tab:flicker_metric}).
Since supersampling only samples at a higher resolution, triangles can still fall between these samples, resulting in flickering.

\Cref{tab:comparison_ms} shows the results under a multi-scale training setup similar to Mip-Splatting~\cite{Yu2024MipSplatting}.
Full resolution is enforced in 30\% of iterations, so almost half use full resolution.
Our method achieves the highest metrics across all but two configurations, where supersampling leads by at most $0.05$\,dB.
Training with our integral thus allows for better convergence compared to $2{\times}2$ supersampling, with differences of up to $\sim$1.9\,dB at $\nicefrac{1}{8}$ on the NeRF Synthetic dataset.
Compared with single-scale training, metrics are lower at the training resolution, since fewer iterations use the full resolution, and higher at $\nicefrac{1}{8}$.

We show the impact of our formulation on rendering speed and how depth-splitting avoids the worst case at low resolutions in \cref{tab:speed}.
The integral costs more per triangle and pixel than a point sample: without the efficiency changes of \cref{sec:efficient}, ours runs at 27--97\,FPS, and with them at 99--322\,FPS.
$2{\times}2$ supersampling of the point-sampled model is faster at every scale (101--350\,FPS with depth split).
Without a depth split, it is even faster than point sampling at $\nicefrac{1}{4}$ and $\nicefrac{1}{8}$ because a low-resolution frame has too few tiles to occupy the GPU, whereas supersampling renders four times as many. 
Depth split removes most of this bottleneck across all renderers.

Lastly, we show how our method compares against Gaussian-based anti-aliasing methods in \cref{tab:comparison_gaussians}.
Without our integral, triangles trail behind when zooming out, with Triangle Splatting$_{\text{no culling}}$ $5.5$\,dB below the best Gaussian method at $\nicefrac{1}{8}$ on Mip-NeRF 360 and $4.6$\,dB below on Tanks and Temples.
Anti-aliasing closes most of this gap, as ours reaches $29.26$ and $24.45$\,dB at $\nicefrac{1}{8}$, ahead of Mip-Splatting and Analytic-Splatting and second only to Gaussian Blending.
$2{\times}2$ supersampling reaches a similar PSNR, but ours has better SSIM and LPIPS at $\nicefrac{1}{8}$, flickers less (\cref{tab:flicker_metric}) and matches or beats it in all but two configurations after multi-scale training (\cref{tab:comparison_ms}).
At the training resolution, ours has the best LPIPS on both data sets and a PSNR within $0.6$\,dB of the best Gaussian method.
After multi-scale training, the best Gaussian method leads in PSNR by $0.3$--$1.5$\,dB at $\nicefrac{1}{8}$, while ours has the best SSIM and LPIPS on Tanks and Temples at $1$ and $\nicefrac{1}{2}$ and improves on Triangle Splatting$_{\text{no culling}}$ by $2.0$--$4.6$\,dB at $\nicefrac{1}{8}$ (supplementary material).
To summarize, our method enables anti-aliasing for Triangle Splatting, allows improved multiscale training, and significantly reduces flickering compared to $2{\times}2$ supersampling.

\begin{table}[b]
\centering\small\setlength{\tabcolsep}{4pt}
\caption{\textbf{Where the remaining error comes from.} Our single-scale Mip-NeRF 360 models rendered with $k{\times}k$ samples per pixel: point samples, blended per pixel, or our exact pixel integral per sub-pixel, blended with one $\alpha$ per triangle and sub-pixel. PSNR (dB) against the $16{\times}16$ point-sampled render of the same model that agrees with $8{\times}8$  to ${>}56$\,dB, indicating convergence. Means over the nine scenes.}
\label{tab:compositing}
\begin{tabular*}{\linewidth}{@{\extracolsep{\fill}}lcccc}
\toprule
& \multicolumn{4}{c}{Point sampling / Ours} \\
\cmidrule(lr){2-5}
Samples & $1$ & $\nicefrac{1}{2}$ & $\nicefrac{1}{4}$ & $\nicefrac{1}{8}$ \\
\midrule
$1{\times}1$ & 42.6\,/\,52.2 & 36.3\,/\,45.7 & 30.2\,/\,39.5 & 26.1\,/\,34.5 \\
$2{\times}2$ & 52.6\,/\,58.8 & 47.8\,/\,54.7 & 42.4\,/\,49.2 & 36.2\,/\,43.3 \\
$4{\times}4$ & 60.1\,/\,63.7 & 56.5\,/\,60.6 & 52.8\,/\,56.8 & 48.2\,/\,52.0 \\
$8{\times}8$ & 66.4\,/\,68.0 & 63.2\,/\,65.1 & 60.0\,/\,61.8 & 56.7\,/\,58.1 \\
\bottomrule
\end{tabular*}
\end{table}

\subsection{Rendering Analysis}
\label{sec:analysis}

\Cref{tab:comparison_ss} raises the question of why $2{\times}2$ supersampling achieves a higher PSNR, even though our integral is exact.
Most rasterizers used for Triangle and Gaussian Splatting make two approximations per pixel: evaluating a primitive at a single point, and assuming opacity is constant within the pixel.
Since our method computes an exact per-primitive integral, we can rule it out as a source of error and focus on blending.
We evaluate this using $16{\times}16$ supersampling as the reference, which agrees with $8{\times}8$ supersampling by over 56\,dB.
As the reference renders the same model, reconstruction errors drop out, the impact of the $\nicefrac{1}{255}$ threshold holds for both, and the remaining difference can be attributed to per-pixel compositing.
\Cref{tab:compositing} shows that, with the same number of samples, ours consistently achieves a higher PSNR.
However, $2{\times}2$ point samples achieve a higher PSNR than our integral with one $\alpha$ per pixel ($36.2$ vs.\ $34.5$\,dB at $\nicefrac{1}{8}$), so compositing one $\alpha$ per pixel introduces more error than sampling with four samples.
Both methods also err in different ways: compositing errs at edges where two triangles share a pixel, and sampling errs on fine texture, appearing as moiré (supplementary material).
This likely explains why supersampling achieves a higher PSNR, while ours achieves better LPIPS and less flickering.
Computing our integral per $2{\times}2$ sub-pixel confirms this, as each $\alpha$ then covers a quarter of the pixel and the PSNR increases from 34.5 to 43.3\,dB.
Across all scales, our integral is $6.2$--$9.6$\,dB ahead of point sampling at the same $1{\times}1$ or $2{\times}2$ samples, whereas with a quarter of the samples it trails point sampling by only $0.4$--$4.9$\,dB (\cref{tab:compositing}).
The exact integral thus recovers most of the gains from using four times as many samples.
The remaining error follows from per-pixel compositing, which Gaussian Blending~\cite{koo2026gb} addresses with opacity distributions, indicating an interesting research direction for future work.

\subsection{Limitations}
While our method computes an exact integral, reduces flickering compared to $2{\times}2$ supersampling, and improves culling over the baseline Triangle Splatting, it comes with limitations.
First, introducing additional computations increases render time relative to baseline Triangle Splatting and even to $2{\times}2$ supersampling.
This gap may be narrowed with further kernel optimizations, as suggested by recent work on 3DGS~\cite{Hahlbohm_2026_CVPR, pajak2026higshierarchicalrenderingarchitecture}.
Second, while our approach allows analysis of the scalar alpha-blending behavior, it does not solve the problem.
Solutions based on A-buffers or the modeling of opacity distributions~\cite{koo2026gb} can be explored.
Our method provides a principled starting point by removing one potential source of error in Triangle Splatting~\cite{held2026triangle}.

\section{Conclusion}
\label{sec:conclusion}

We derived an exact closed-form integral of the Triangle Splatting window over a pixel, together with its gradients.
Integrating rather than point sampling significantly improves multi-scale training, especially at low resolutions, and produces more temporally stable renderings, with considerably less flickering than both point sampling and $2{\times}2$ supersampling.
The same formulation allows us to replace the inradius cull with a conservative, integral-based cull that removes only triangles that cannot affect any pixel.
Finally, because the integral is exact, it can be ruled out as a source of error, so future work can focus on analyzing the remaining sources of error.

\newpage
\section*{Acknowledgements}

This research was partly funded by the FWO fellowship grants (1SHDZ24N, 1S80926N), the Special Research Fund (BOF) of Hasselt University (R-14360, R-14436), the Flanders Make SBO projects NORM.AI (R-13936), and ALARMM (R-15482). This work was made possible by MAXVR-INFRA, a scalable and flexible infrastructure that facilitates the transition to digital-physical work environments.

{
    \small
    \bibliographystyle{ieeenat_fullname}
    \bibliography{main}
}
\clearpage
\setcounter{page}{1}
\maketitlesupplementary
This supplement extends the experiments of the main paper and provides additional derivations of \cref{sec:exact_integral,sec:backward}.
The supplementary video first zooms out on three scenes from the training resolution to $\nicefrac{1}{8}$, comparing Triangle Splatting$_{\text{no culling}}$ with ours, and then shows the sub-pixel pan of \cref{sec:supp:flicker}.

\section{Temporal Stability}
\label{sec:supp:flicker}

\begin{figure*}[t]
  \centering
  \includegraphics[width=1.0\textwidth]{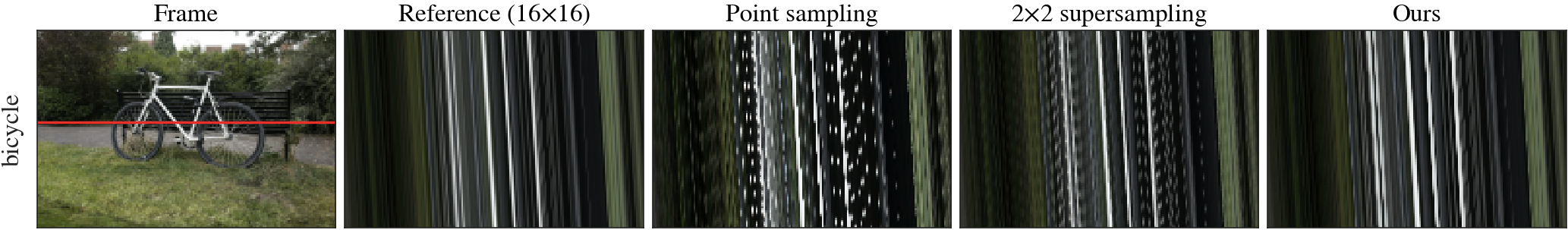}
  \includegraphics[width=1.0\textwidth]{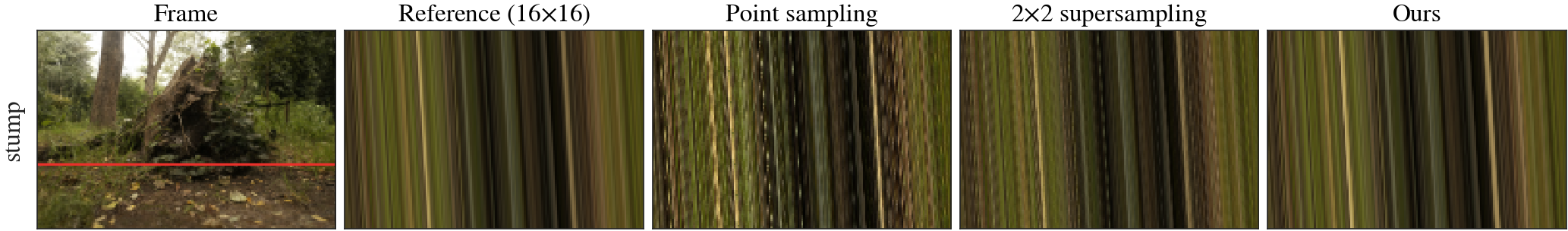}
  \includegraphics[width=1.0\textwidth]{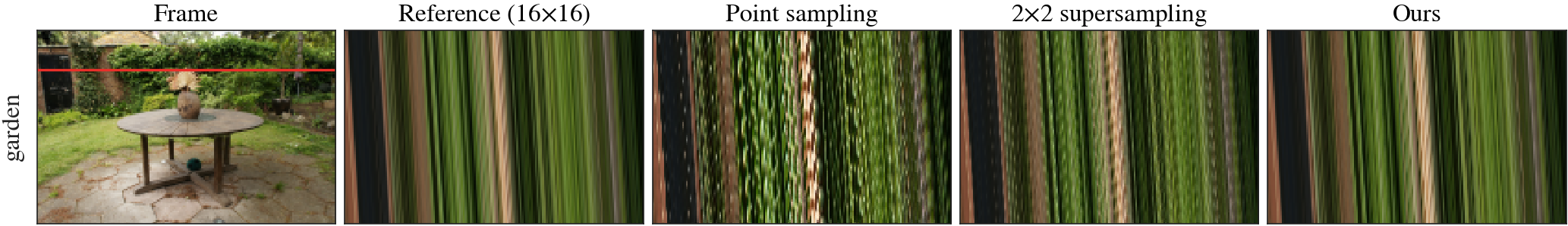}
  \includegraphics[width=1.0\textwidth]{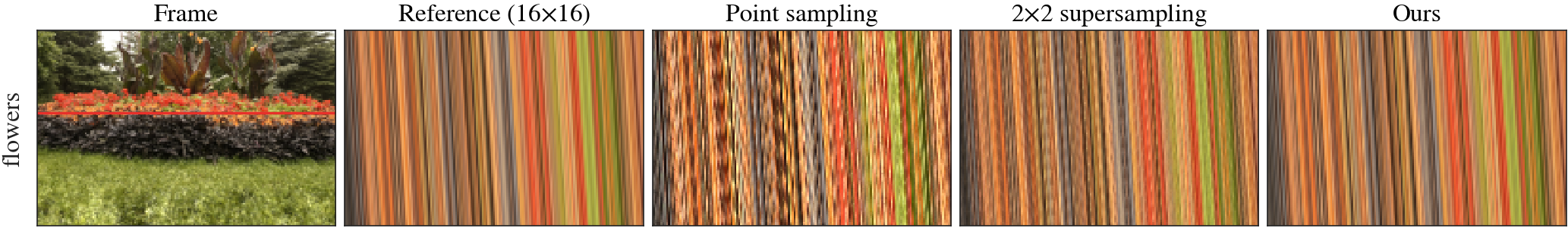}
  \caption{\textbf{Spatio-temporal slices} of a sub-pixel pan at $\nicefrac{1}{8}$ of the training resolution. The camera moves by $\nicefrac{1}{16}$ pixel per frame, and each slice stacks the marked image row over the frames. All renderers draw the same model using Triangle Splatting$_{\text{no culling}}$. The reference renders it with $16{\times}16$ supersampling, since no captured images exist for the pan. Point sampling breaks thin structures into dashes as they enter and leave the sample positions, $2{\times}2$ supersampling reduces this, and ours stays closest to the reference.}
  \label{fig:spatio_temporal_slices}
\end{figure*}

We measure flickering on a pan that moves the image by $\nicefrac{1}{16}$ pixel per frame over $T = 128$ frames.
With $X_t$ the rendered frame $t$ and $R_t$ the $16{\times}16$ supersampled render of the same model, the error of the frame-to-frame differences in \cref{tab:flicker_metric} is
\begin{equation}
  \sqrt{\frac{1}{(T-1)\,N} \sum_{t=1}^{T-1} \big\| (X_{t+1} - X_t) - (R_{t+1} - R_t) \big\|^2}
  \label{eq:supp_flicker}
\end{equation}
where the norm runs over the pixels and color channels, and $N$ is the number of them.
Comparing differences between frames rather than the frames themselves removes errors that stay the same in every frame, so only changes from frame to frame are counted.
\Cref{fig:spatio_temporal_slices} shows spatio-temporal slices of the pan, and the supplementary video shows the same pan for three scenes.
Point sampling breaks thin structures into dashes as sub-pixel triangles enter and leave the sample positions.
$2{\times}2$ supersampling reduces this, and our integral changes continuously with the camera and stays closest to the reference.

\begin{table}[b]
\centering
\caption{\textbf{Drop-in renderer.} PSNR (dB) on Mip-NeRF~360 of the single-scale Triangle Splatting$_{\text{original}}$ models, trained with point sampling and the inradius cull~\cite{held2026triangle}, rendered without retraining.}
\label{tab:dropin}
\small\setlength{\tabcolsep}{3pt}
\begin{tabular*}{\linewidth}{@{\extracolsep{\fill}}lcccc@{}}
\toprule
Rendered with & $1$ & $\nicefrac{1}{2}$ & $\nicefrac{1}{4}$ & $\nicefrac{1}{8}$ \\
\midrule
Point sampling, with cull    & 27.18 & 26.76 & 20.13 & 12.93 \\
Point sampling, without cull & 27.04 & 27.65 & 26.83 & 24.86 \\
$2{\times}2$ supersampling   & 27.18 & 28.32 & 29.59 & 29.86 \\
Ours                         & 27.18 & 28.25 & 29.19 & 29.28 \\
\bottomrule
\end{tabular*}
\end{table}

\section{Drop-in Renderer}
\label{sec:supp:dropin}

\Cref{tab:dropin} renders single-scale Triangle Splatting models trained with point sampling and the inradius cull, as released~\cite{held2026triangle}, without retraining.
The inradius cull removes every triangle whose inradius is below one pixel, and zooming out shrinks every inradius, so point sampling with the cull drops to $12.93$\,dB at $\nicefrac{1}{8}$, against $24.86$\,dB without it.
Our integral and 2$\times$2 supersampling raise this to $29.28$ and $29.86$\,dB.

\begin{figure}[t]
  \centering
  \includegraphics[width=1.0\columnwidth]{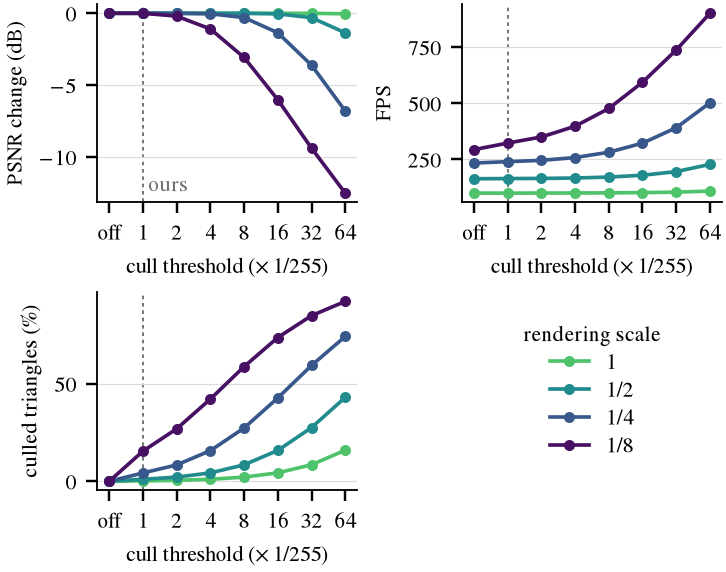}
  \caption{\textbf{Culling threshold sweep} on the single-scale Mip-NeRF~360 models, with depth split, mean over the nine scenes. The threshold is given in multiples of the alpha cutoff $\nicefrac{1}{255}$, \emph{off} disables the cull, and the dashed line marks ours. Top left: PSNR change against no culling. Top right: frame rate. Bottom left: share of culled triangles.}
  \label{fig:culling_sweep}
\end{figure}

\section{Culling Parameter Sweep}
\label{sec:supp:cull}

Our cull removes a triangle only if it stays below the alpha cutoff at every pixel, where the renderer skips it anyway, so, in exact arithmetic, it changes no pixels.
Rendering our single-scale Mip-NeRF 360 models with and without it, the images agree to at least $107$\,dB PSNR, and the largest per-pixel difference, $0.002$, arises from floating-point rounding at the threshold.
The cull removes $0.2\%$, $1.0\%$, $4.3\%$ and $15.4\%$ of the triangles at $1$, $\nicefrac{1}{2}$, $\nicefrac{1}{4}$ and $\nicefrac{1}{8}$, and raises the frame rate at $\nicefrac{1}{8}$ by $10\%$ with depth split and by $21\%$ without.
\Cref{fig:culling_sweep} raises the threshold above the alpha cutoff.
At high resolution, a larger threshold barely changes the number of culled triangles.
At low resolution, however, it culls over $75\%$ of the triangles and lowers PSNR by up to ${\sim}12$\,dB.
The frame rate rises as fewer triangles remain.

\section{Changes to Triangle Splatting}
\label{sec:supp_clone}

\paragraph{Cloning.}
When Triangle Splatting clones a triangle, both copies receive the opacity
\begin{equation}
  o' = 1 - \sqrt{1 - o}, \qquad o = \operatorname{sigmoid}(l),
  \label{eq:supp_clone}
\end{equation}
so that the two copies together cover a pixel as much as the original, and the new logit is $l' = \operatorname{logit}(o')$.
The released code evaluates this using single-precision opacity.
For logits above about $17$, $\operatorname{sigmoid}(l)$ rounds to exactly $1$, so $o' = 1$ and $l' = \infty$.
Such a triangle has zero gradient with respect to its opacity and can never become transparent again.
Since triangles are selected for cloning by their opacity, these copies continue to multiply.
The released Triangle Splatting models reach these logits after single-scale training on 6 of our 13 scenes, and after multi-scale training more often.
We instead evaluate \cref{eq:supp_clone} directly in logits.
With $1 - \operatorname{sigmoid}(l) = \operatorname{sigmoid}(-l)$,
\begin{equation}
  q = \sqrt{\operatorname{sigmoid}(-l)}, \qquad l' = \ln \frac{1 - q}{q},
  \label{eq:supp_clone_logit}
\end{equation}
which is mathematically identical to \cref{eq:supp_clone}, matches a double-precision evaluation and stays finite.
We use it for every triangle model we train, including Triangle Splatting$_{\text{original}}$.

\paragraph{Spherical harmonics.}
The released Triangle Splatting implementation passes the triangle center to the spherical harmonics (SH) evaluation as \texttt{(glm::vec3)(x, y, z)}, which the comma operator reduces to the point $(z, z, z)$.
The view direction of the SH therefore points to $(z, z, z)$ instead of the center of the triangle.
This issue was pointed out on GitHub and has since been fixed in MeshSplatting~\cite{Held_2026_CVPR}.
All triangle models in the paper retain the released SH evaluation and change only the cloning step, since fixing it lowered the PSNR for baseline Triangle Splatting in our tests.

\section{Derivations}
\label{sec:supp:derivations}

The forward integral is derived in full in \cref{sec:exact_integral}.
Here we provide the Taylor expansion used when $u$ changes little along a side, show that the moving limits add no terms to the gradients in \cref{sec:backward}, and derive the closed-form side terms of the gradients.

\subsection{Taylor Expansion}
\label{sec:supp:taylor}

For sides with $u_A \approx u_B$, the numerator and denominator of the quotient in \cref{eq:almost_final} both vanish, and single precision loses most digits to cancellation.
When $|u_B - u_A| \leq 0.05 \max(u_A, u_B)$, we expand around $m = (u_A + u_B)/2$.
With $\epsilon = (u_B - u_A)/m$, the odd terms cancel and
\begin{equation}
  \begin{aligned}
    \frac{u_B^{\sigma+2} - u_A^{\sigma+2}}{(\sigma+2)(u_B - u_A)}
    = m^{\sigma+1}\Big(1 &+ \tfrac{(\sigma+1)\sigma}{24}\,\epsilon^2 \\
    &+ \tfrac{(\sigma+1)\sigma(\sigma-1)(\sigma-2)}{1920}\,\epsilon^4\Big),
  \end{aligned}
  \label{eq:supp_taylor}
\end{equation}
up to and including the term of order $\epsilon^4$.
The side integrals $U$, $V$ and $G$ of \cref{sec:supp:gradients} are expanded the same way.

\subsection{Variable Limits Leibniz Simplification}
\label{sec:supp:limits}

To apply the Leibniz rule with variable limits, the derivative can be pushed inside. 
This however requires extra terms in front for which we show they cancel, resulting in a simpler formulation.
Let $\theta$ be one of the $\mathbf{n}$ and $O$ of an edge.
We write $C$ as an integral over vertical columns of the pixel, each running from the pixel's bottom $y_\text{bottom}$ to its top $y_\text{top}$, with $I = 0$ outside the triangle.
A column is cut into segments where it crosses a triangle edge or a bisector, and on each segment the integrand is $u_k^\sigma$ for one sub-triangle $W_k$, or $0$, where $u_k$ is $u$ of \cref{eq:eq_u} for edge $k$.
For example, a column that enters the triangle through edge $i$, crosses the bisector into $W_j$, and leaves through edge $j$ gives
\begin{equation}
  \begin{aligned}
    \int_{y_\text{bottom}}^{y_\text{top}} I\, dy
    &= \int_{y_\text{bottom}}^{y_{\text{edge}\,i}} 0\, dy
     + \int_{y_{\text{edge}\,i}}^{y_\text{bisector}} u_i^{\sigma}\, dy \\
    &\quad + \int_{y_\text{bisector}}^{y_{\text{edge}\,j}} u_j^{\sigma}\, dy
     + \int_{y_{\text{edge}\,j}}^{y_\text{top}} 0\, dy .
  \end{aligned}
  \label{eq:supp_column}
\end{equation}
The Leibniz rule with variable limits, applied to each segment, adds the integrand at each limit times the rate of change of that limit.
Grouped by limit, these terms are
\begin{equation}
  \begin{aligned}
    &- u_i^{\sigma}\big|_{y_{\text{edge}\,i}} \frac{d y_{\text{edge}\,i}}{d\theta}
     + \big[ u_i^{\sigma} - u_j^{\sigma} \big]_{y_\text{bisector}} \frac{d y_\text{bisector}}{d\theta} \\
    &+ u_j^{\sigma}\big|_{y_{\text{edge}\,j}} \frac{d y_{\text{edge}\,j}}{d\theta}
     + 0 \cdot \frac{d y_\text{top}}{d\theta} - 0 \cdot \frac{d y_\text{bottom}}{d\theta} .
  \end{aligned}
  \label{eq:supp_leibniz}
\end{equation}
The first and third terms are zero because $u_i = 0$ on edge $i$ and $u_j = 0$ on edge $j$, so the window is $0^\sigma = 0$ there.
The second is zero because $u_i = u_j$ on the bisector.
The last two are zero because the window is $0$ outside the triangle and the pixel border does not move.
The same holds for every column, and the columns span the pixel's width, which does not move either, so
$\frac{d}{d\theta} \iint I\, dx\, dy = \sum_W \iint_D \frac{\partial I}{\partial \theta}\, dx\, dy$,
which gives \cref{eq:grad_area}.

\subsection{Gradients}
\label{sec:supp:gradients}

\paragraph{Inradius gradient.}
The gradients of \cref{sec:backward} hold the inradius $r$ fixed, as Triangle Splatting does.
Including it adds $\partial C_W/\partial r = -\sigma C_W / r$, chained to the vertices through $r = 2A/(L_0 + L_1 + L_2)$.
We retrained the nine single-scale Mip-NeRF 360 models with this term.
On the eight that converged, the mean PSNR changes by $-0.03$, $-0.03$, $-0.03$ and $+0.08$\,dB at $1$, $\nicefrac{1}{2}$, $\nicefrac{1}{4}$ and $\nicefrac{1}{8}$, within the variation between training runs.
On \emph{counter}, training diverged, since the term grows as $1/r$ for the sub-pixel triangles that our cull keeps.
We therefore keep $r$ fixed.

\paragraph{Derivatives of the window.}
Inside sub-triangle $W$ the window is $I = u^\sigma$, and only $u$ depends on $\mathbf{n}$ and $O$.
With $r$ held fixed, \cref{eq:eq_u} gives $\partial u / \partial O = -1/r$ and $\partial u / \partial \mathbf{n} = -\mathbf{p}/r$, so
\begin{equation}
  \begin{aligned}
    \frac{\partial I}{\partial O} &= -\frac{\sigma}{r} u^{\sigma-1}, \qquad
    \frac{\partial I}{\partial \mathbf{n}}
      = -\frac{\sigma}{r} u^{\sigma-1} \mathbf{p}, \\
    \frac{\partial I}{\partial \sigma} &= u^{\sigma} \ln u ,
  \end{aligned}
  \label{eq:supp_dI}
\end{equation}
and integrating these over $D$ gives \cref{eq:grad_area}.
We solve each of the three integrals the same way as $C$ in \cref{sec:exact_integral}.
We change the variables to $u$ and $a$, so that $dx\,dy = r\,du\,da$, apply Green's theorem (\cref{eq:green}) with $Q = 0$ and a $P$ chosen for the integrand, and parameterize each side of $E$ from $A$ to $B$ with $\lambda \in [0, 1]$ as in \cref{eq:parameterization}:
\begin{equation}
  \begin{aligned}
    u_\lambda &= u_A + \lambda (u_B - u_A), \\
    a_\lambda &= a_A + \lambda (a_B - a_A), \\
    da &= (a_B - a_A)\, d\lambda .
  \end{aligned}
  \label{eq:supp_lambda}
\end{equation}

\paragraph{Offset.}
Changing the variables in \cref{eq:grad_area} gives
\begin{equation}
  \frac{\partial C_W}{\partial O}
    = -\sigma \iint_{D'} u^{\sigma-1}\, du\, da ,
  \label{eq:supp_dO_area}
\end{equation}
which is the integral of $C$ with the exponent $\sigma - 1$ instead of $\sigma$.
We set $Q = 0$, so that $-\frac{\partial P}{\partial u} = u^{\sigma-1}$ must hold, which gives $P = -u^{\sigma}/\sigma$.
Green's theorem turns the area integral into a sum over the sides
\begin{equation}
  \frac{\partial C_W}{\partial O}
    = -\sigma \oint_E -\frac{u^\sigma}{\sigma}\, da
    = \sum_E \int_{a_A}^{a_B} u^{\sigma}\, da .
  \label{eq:supp_dO_line}
\end{equation}
Substituting \cref{eq:supp_lambda} gives
\begin{equation}
  \frac{\partial C_W}{\partial O}
    = \sum_E (a_B - a_A) \int_0^1 u_\lambda^{\sigma}\, d\lambda
    = -\sum_E (a_A - a_B)\, U ,
  \label{eq:supp_dO}
\end{equation}
with $U = \int_0^1 u_\lambda^{\sigma}\, d\lambda$ per side.

\paragraph{Normal.}
Changing the variables in \cref{eq:grad_area} gives
\begin{equation}
  \frac{\partial C_W}{\partial \mathbf{n}}
    = -\sigma \iint_{D'} \mathbf{p}\, u^{\sigma-1}\, du\, da .
  \label{eq:supp_dn_area}
\end{equation}
Here $\mathbf{p}$ also changes over $D'$, so we first write it in terms of $u$ and $a$.
Since $\mathbf{n}$ and $\mathbf{t}$ are orthonormal, $\mathbf{p} = (\mathbf{t} \cdot \mathbf{p})\, \mathbf{t} + (\mathbf{n} \cdot \mathbf{p})\, \mathbf{n}$, with $\mathbf{t} \cdot \mathbf{p} = a$ from \cref{eq:a_transform} and $\mathbf{n} \cdot \mathbf{p} = -r u - O$ from \cref{eq:eq_u}:
\begin{equation}
  \mathbf{p} = a\, \mathbf{t} - (r u + O)\, \mathbf{n} .
  \label{eq:supp_p}
\end{equation}
Substituting into \cref{eq:supp_dn_area} splits it into three integrals
\begin{equation}
  \begin{aligned}
    \frac{\partial C_W}{\partial \mathbf{n}}
      = &-\sigma\, \mathbf{t} \iint_{D'} a\, u^{\sigma-1}\, du\, da \\
        &+ \sigma\, \mathbf{n}\, r \iint_{D'} u^{\sigma}\, du\, da \\
        &+ \sigma\, \mathbf{n}\, O \iint_{D'} u^{\sigma-1}\, du\, da .
  \end{aligned}
  \label{eq:supp_dn_split}
\end{equation}
The second line is $\sigma C_W \mathbf{n}$ by \cref{eq:dadu_surface}, and the third is $-O \frac{\partial C_W}{\partial O} \mathbf{n}$ by \cref{eq:supp_dO_area}.
For the first, we set $Q = 0$ and $P = -a u^{\sigma}/\sigma$, so that $-\frac{\partial P}{\partial u} = a u^{\sigma-1}$.
Green's theorem and \cref{eq:supp_lambda} give
\begin{equation}
  \begin{aligned}
    \iint_{D'} a\, u^{\sigma-1}\, du\, da
      &= -\frac{1}{\sigma} \sum_E \int_{a_A}^{a_B} a\, u^{\sigma}\, da \\
      &= -\frac{1}{\sigma} \sum_E (a_B - a_A)
         \int_0^1 a_\lambda u_\lambda^{\sigma}\, d\lambda .
  \end{aligned}
  \label{eq:supp_a_line}
\end{equation}
Since $a_\lambda = a_A + \lambda (a_B - a_A)$, the integral along a side is $a_A U + (a_B - a_A) V$, with $V = \int_0^1 \lambda\, u_\lambda^{\sigma}\, d\lambda$.
Putting the three lines together gives
\begin{equation}
  \begin{aligned}
    \frac{\partial C_W}{\partial \mathbf{n}}
      = &-\Big[ \sum_E (a_A - a_B)
          \big( a_A U + (a_B - a_A) V \big) \Big]\, \mathbf{t} \\
        &+ \Big( \sigma C_W - O \frac{\partial C_W}{\partial O} \Big)\,
          \mathbf{n} .
  \end{aligned}
  \label{eq:supp_dn}
\end{equation}
The part along $\mathbf{n}$ only reuses $C_W$ and $\partial C_W / \partial O$, so the sides only need $U$ and $V$.

\paragraph{Smoothness.}
Changing the variables in \cref{eq:grad_area} gives
\begin{equation}
  \frac{\partial C_W}{\partial \sigma}
    = r \iint_{D'} u^{\sigma} \ln u\, du\, da .
  \label{eq:supp_ds_area}
\end{equation}
We set $Q = 0$ and $P = -h(u)$, with $h$ an antiderivative of $u^{\sigma} \ln u$.
Integration by parts gives
\begin{equation}
  h(u) = \frac{u^{\sigma+1} \ln u}{\sigma+1}
       - \frac{u^{\sigma+1}}{(\sigma+1)^2} .
  \label{eq:supp_h}
\end{equation}
Green's theorem and \cref{eq:supp_lambda} then give
\begin{equation}
  \begin{aligned}
    \frac{\partial C_W}{\partial \sigma}
      = &-\frac{r}{\sigma+1} \sum_E (a_B - a_A)
         \int_0^1 u_\lambda^{\sigma+1} \ln u_\lambda\, d\lambda \\
        &+ \frac{r}{(\sigma+1)^2} \sum_E (a_B - a_A)
         \int_0^1 u_\lambda^{\sigma+1}\, d\lambda .
  \end{aligned}
  \label{eq:supp_ds_split}
\end{equation}
The integral in the second line is the quotient in \cref{eq:almost_final}, as $du = (u_B - u_A)\, d\lambda$ gives
\begin{equation}
  \begin{aligned}
    \int_0^1 u_\lambda^{\sigma+1}\, d\lambda
      &= \frac{1}{u_B - u_A} \int_{u_A}^{u_B} u^{\sigma+1}\, du \\
      &= \frac{u_B^{\sigma+2} - u_A^{\sigma+2}}{(\sigma+2)(u_B - u_A)} ,
  \end{aligned}
  \label{eq:supp_quot}
\end{equation}
so the second line is $-C_W / (\sigma + 1)$, and
\begin{equation}
  \frac{\partial C_W}{\partial \sigma}
    = \frac{r}{\sigma+1} \sum_E (a_A - a_B)\, G
      - \frac{C_W}{\sigma+1} ,
  \label{eq:supp_ds}
\end{equation}
with $G = \int_0^1 u_\lambda^{\sigma+1} \ln u_\lambda\, d\lambda$ per side.
As $D$ does not depend on $\sigma$, this is also the derivative of \cref{eq:almost_final} with respect to $\sigma$.

\paragraph{Side integrals.}
We solve $U$ and $G$ with the same substitution $du = (u_B - u_A)\, d\lambda$ as in \cref{eq:supp_quot}, using $h$ from \cref{eq:supp_h} with $\sigma + 1$ in place of $\sigma$ for $G$:
\begin{equation}
  \begin{aligned}
    U &= \frac{1}{u_B - u_A} \int_{u_A}^{u_B} u^{\sigma}\, du
       = \frac{u_B^{\sigma+1} - u_A^{\sigma+1}}{(\sigma+1)(u_B - u_A)} , \\
    G &= \frac{1}{u_B - u_A} \int_{u_A}^{u_B} u^{\sigma+1} \ln u\, du \\
      &= \frac{1}{u_B - u_A}
         \Big[ \frac{v^{\sigma+2} \ln v}{\sigma+2}
             - \frac{v^{\sigma+2}}{(\sigma+2)^2} \Big]_{v=u_A}^{u_B} .
  \end{aligned}
  \label{eq:supp_UG}
\end{equation}
For $V$ we integrate by parts in $\lambda$, with $u_\lambda^{\sigma} = \frac{d}{d\lambda} \frac{u_\lambda^{\sigma+1}}{(\sigma+1)(u_B - u_A)}$, and use \cref{eq:supp_quot} for the remaining integral:
\begin{equation}
  \begin{aligned}
    V &= \Big[ \frac{\lambda\, u_\lambda^{\sigma+1}}
               {(\sigma+1)(u_B - u_A)} \Big]_0^1
       - \frac{\int_0^1 u_\lambda^{\sigma+1}\, d\lambda}
              {(\sigma+1)(u_B - u_A)} \\
      &= \frac{u_B^{\sigma+1}}{(\sigma+1)(u_B - u_A)}
       - \frac{u_B^{\sigma+2} - u_A^{\sigma+2}}
              {(\sigma+1)(\sigma+2)(u_B - u_A)^2} .
  \end{aligned}
  \label{eq:supp_V}
\end{equation}
When $u_A = u_B$, $u_\lambda = u_A$ along the whole side, so $U = u_A^{\sigma}$, $V = \frac{1}{2} u_A^{\sigma}$ and $G = u_A^{\sigma+1} \ln u_A$.
On the triangle edge $u = 0$, where we use the limit $v^{\sigma+2} \ln v = 0$.
For $\sigma < 1$ the integrand $u^{\sigma-1}$ in \cref{eq:supp_dO_area} is unbounded at the edge, but $U$ and $V$ only contain $u^{\sigma+1}$ and $u^{\sigma+2}$ and stay finite.

Each gradient is thus a closed-form term on each side of $D$, like $C$ itself, and a single pass over the sides yields $C$ and all its gradients.
A bisector side belongs to both sub-triangles next to it, each with its own $\mathbf{n}$, $O$ and $u$, so it sends gradients to both edges.

\section{Additional Results}
\label{sec:supp:results}

\begin{figure*}[t]
  \centering
  \includegraphics[width=1.0\linewidth]{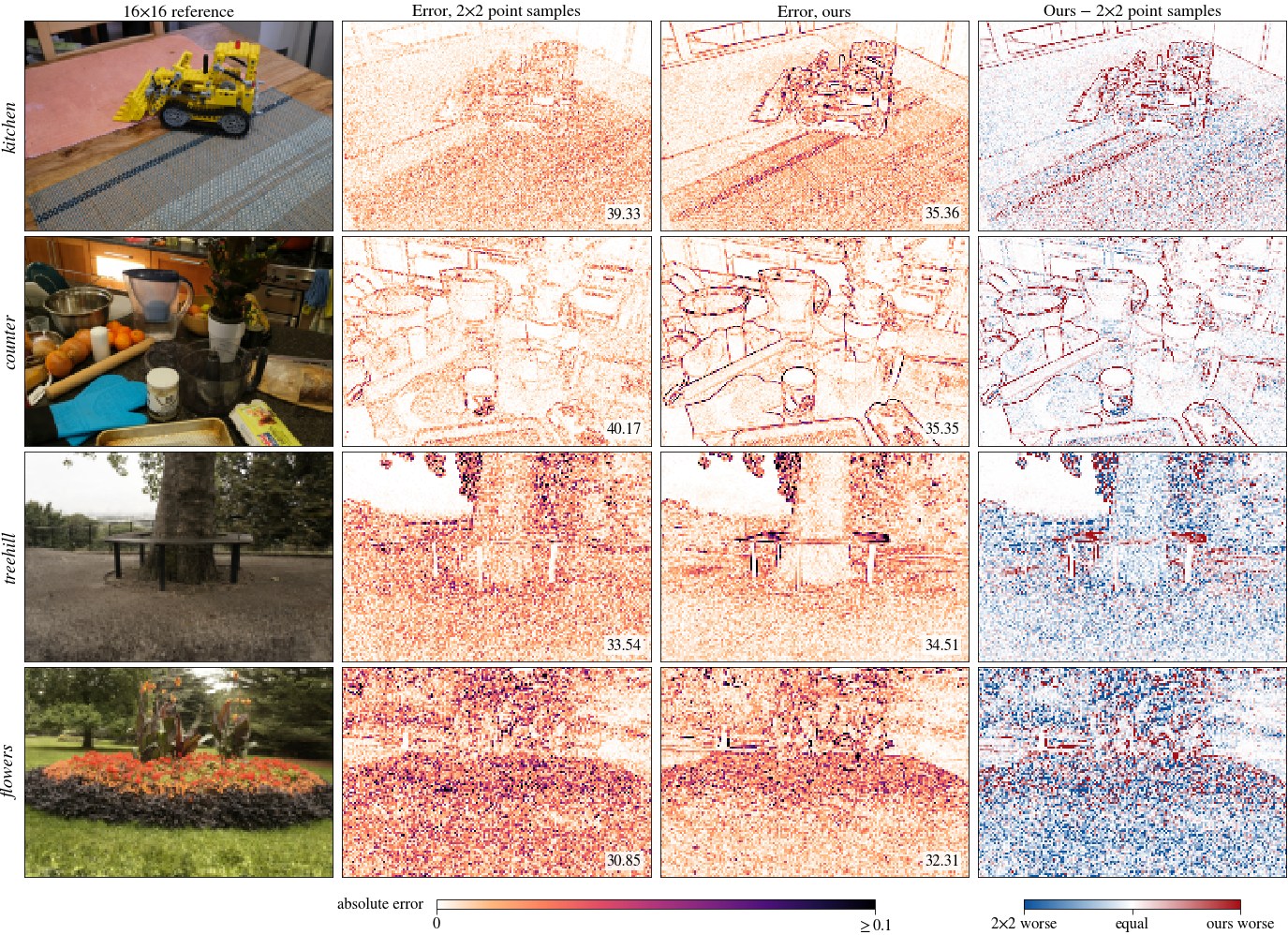}
  \caption{\textbf{Error comparison.}
    We render our single-scale Mip-NeRF 360 models at $\nicefrac{1}{8}$ of the training resolution and measure the error against a $16{\times}16$ point-sampled render of the same model, as in \cref{tab:compositing}. The second and third columns show the per-pixel absolute error of $2{\times}2$ point samples and of ours with one $\alpha$ per pixel, with their PSNR in dB. The last column shows which method has the higher error, red for ours and blue for $2{\times}2$ point samples. Our errors concentrate on edges, whereas point sampling mainly errs on textured areas, which agrees with per-pixel compositing being the main source of error in our method.}
  \label{fig:error_comp}
\end{figure*}

\Cref{tab:supp:full_ss,tab:supp:full_ms} extend \cref{tab:comparison_ss,tab:comparison_ms,tab:comparison_gaussians} to every method, scale, and dataset, including Triangle Splatting$_{\text{original}}$ with the inradius cull.
Additionally, \cref{fig:error_comp} shows where the per-pixel errors in \cref{tab:compositing} occur across four Mip-NeRF 360 scenes.

\begin{table*}[t]
\centering\small\setlength{\tabcolsep}{3pt}
\caption{\textbf{Zoom-out after single-scale training} on Mip-NeRF 360~\cite{barron2022mipnerf360}, Tanks and Temples~\cite{Knapitsch2017} (truck, train), Deep Blending~\cite{deep_blending} (drjohnson, playroom) and NeRF Synthetic~\cite{mildenhall2020nerf}, for every method. Test views are rendered at $1$ to $\nicefrac{1}{8}$ of the training resolution and compared with the box-downsampled ground truth. We report the mean across scenes for each data set. Triangle Splatting$_{\text{original}}$ point-samples and culls sub-pixel triangles, as released. Triangle Splatting$_{\text{no culling}}$ keeps them in training and rendering, and Triangle Splatting$_{\text{supersampled}}$ renders that model with $2{\times}2$ samples per pixel. The triangle models are trained in our framework using the released settings and the stable opacity update from \cref{sec:supp_clone}, while the Gaussian methods are trained using the authors' code and settings. All methods are evaluated with the same protocol. \colorbox{tabfirstc}{Blue} marks the best value.}
\label{tab:supp:full_ss}
\resizebox{\textwidth}{!}{%
\begin{tabular}{lcccccccccccc}
\toprule
Method & \multicolumn{3}{c}{$1$} & \multicolumn{3}{c}{$\nicefrac{1}{2}$} & \multicolumn{3}{c}{$\nicefrac{1}{4}$} & \multicolumn{3}{c}{$\nicefrac{1}{8}$} \\
\cmidrule(lr){2-4}\cmidrule(lr){5-7}\cmidrule(lr){8-10}\cmidrule(lr){11-13}
 & PSNR$^{\uparrow}$ & SSIM$^{\uparrow}$ & LPIPS$^{\downarrow}$ & PSNR$^{\uparrow}$ & SSIM$^{\uparrow}$ & LPIPS$^{\downarrow}$ & PSNR$^{\uparrow}$ & SSIM$^{\uparrow}$ & LPIPS$^{\downarrow}$ & PSNR$^{\uparrow}$ & SSIM$^{\uparrow}$ & LPIPS$^{\downarrow}$ \\
\midrule
\multicolumn{13}{l}{\emph{Mip-NeRF 360}} \\
3DGS~\cite{kerbl3Dgaussians} & 27.52 & 0.814 & 0.217 & 25.09 & 0.807 & 0.180 & 21.29 & 0.690 & 0.208 & 18.79 & 0.567 & 0.240 \\
Mip-Splatting~\cite{Yu2024MipSplatting} & \tabfirst{27.66} & \tabfirst{0.827} & 0.190 & \tabfirst{28.81} & \tabfirst{0.877} & \tabfirst{0.110} & 29.54 & \tabfirst{0.913} & \tabfirst{0.070} & 28.89 & 0.925 & \tabfirst{0.060} \\
Analytic-Splatting~\cite{10.1007/978-3-031-72643-9_17} & 27.52 & 0.816 & 0.215 & 28.64 & 0.866 & 0.137 & 29.18 & 0.903 & 0.091 & 28.47 & 0.917 & 0.074 \\
Gaussian Blending~\cite{koo2026gb} & 26.86 & 0.803 & 0.237 & 27.99 & 0.853 & 0.161 & 29.36 & 0.900 & 0.104 & \tabfirst{30.37} & \tabfirst{0.931} & 0.068 \\
\cmidrule(lr){1-13}
Triangle Splatting$_{\text{original}}$ & 27.18 & 0.814 & 0.191 & 26.76 & 0.817 & 0.148 & 20.13 & 0.593 & 0.281 & 12.93 & 0.251 & 0.581 \\
Triangle Splatting$_{\text{no culling}}$ & 27.16 & 0.815 & 0.190 & 27.78 & 0.850 & 0.121 & 26.87 & 0.827 & 0.130 & 24.83 & 0.765 & 0.178 \\
Triangle Splatting$_{\text{supersampled}}$ & 27.27 & 0.819 & 0.192 & 28.41 & 0.868 & 0.113 & \tabfirst{29.69} & 0.904 & 0.072 & 29.90 & 0.908 & 0.072 \\
\cmidrule(lr){1-13}
Ours & 27.26 & 0.819 & \tabfirst{0.187} & 28.36 & 0.866 & 0.113 & 29.33 & 0.902 & 0.075 & 29.26 & 0.920 & 0.062 \\
\midrule
\multicolumn{13}{l}{\emph{Tanks and Temples}} \\
3DGS~\cite{kerbl3Dgaussians} & \tabfirst{23.87} & 0.851 & 0.170 & 22.35 & 0.845 & 0.130 & 19.40 & 0.744 & 0.155 & 17.10 & 0.619 & 0.207 \\
Mip-Splatting~\cite{Yu2024MipSplatting} & 23.79 & 0.858 & 0.157 & \tabfirst{24.56} & 0.899 & 0.090 & 24.76 & 0.919 & 0.063 & 24.07 & 0.919 & 0.062 \\
Analytic-Splatting~\cite{10.1007/978-3-031-72643-9_17} & 23.75 & 0.853 & 0.173 & 24.53 & 0.894 & 0.104 & 24.62 & 0.914 & 0.070 & 23.78 & 0.912 & 0.067 \\
Gaussian Blending~\cite{koo2026gb} & 23.27 & 0.846 & 0.188 & 24.11 & 0.888 & 0.117 & \tabfirst{24.97} & 0.919 & 0.070 & \tabfirst{25.41} & \tabfirst{0.936} & \tabfirst{0.053} \\
\cmidrule(lr){1-13}
Triangle Splatting$_{\text{original}}$ & 23.11 & 0.857 & 0.143 & 22.26 & 0.845 & 0.130 & 15.95 & 0.596 & 0.309 & 7.93 & 0.184 & 0.614 \\
Triangle Splatting$_{\text{no culling}}$ & 23.14 & 0.860 & 0.140 & 23.54 & 0.888 & 0.092 & 22.57 & 0.864 & 0.115 & 20.83 & 0.823 & 0.156 \\
Triangle Splatting$_{\text{supersampled}}$ & 23.24 & 0.863 & 0.142 & 24.02 & \tabfirst{0.903} & \tabfirst{0.081} & 24.78 & \tabfirst{0.928} & \tabfirst{0.057} & 24.58 & 0.926 & 0.066 \\
\cmidrule(lr){1-13}
Ours & 23.30 & \tabfirst{0.864} & \tabfirst{0.136} & 24.02 & \tabfirst{0.903} & \tabfirst{0.081} & 24.56 & 0.925 & 0.059 & 24.45 & 0.931 & 0.056 \\
\midrule
\multicolumn{13}{l}{\emph{Deep Blending}} \\
3DGS~\cite{kerbl3Dgaussians} & \tabfirst{29.82} & 0.903 & \tabfirst{0.244} & 29.54 & 0.910 & 0.158 & 27.51 & 0.885 & 0.112 & 24.66 & 0.815 & 0.137 \\
Mip-Splatting~\cite{Yu2024MipSplatting} & 29.31 & 0.899 & \tabfirst{0.244} & 29.66 & 0.913 & \tabfirst{0.150} & 29.96 & 0.927 & 0.081 & 29.47 & 0.929 & 0.067 \\
Analytic-Splatting~\cite{10.1007/978-3-031-72643-9_17} & 29.60 & 0.905 & 0.247 & 29.94 & \tabfirst{0.917} & 0.153 & 30.21 & 0.931 & \tabfirst{0.079} & 29.61 & 0.932 & 0.064 \\
Gaussian Blending~\cite{koo2026gb} & 29.61 & \tabfirst{0.906} & 0.258 & \tabfirst{29.98} & 0.916 & 0.171 & \tabfirst{30.63} & \tabfirst{0.932} & 0.093 & \tabfirst{31.39} & \tabfirst{0.945} & \tabfirst{0.062} \\
\cmidrule(lr){1-13}
Triangle Splatting$_{\text{original}}$ & 28.92 & 0.892 & 0.246 & 29.01 & 0.892 & 0.175 & 27.17 & 0.853 & 0.187 & 20.02 & 0.613 & 0.398 \\
Triangle Splatting$_{\text{no culling}}$ & 28.94 & 0.893 & 0.246 & 29.19 & 0.897 & 0.167 & 29.27 & 0.901 & 0.128 & 28.28 & 0.891 & 0.136 \\
Triangle Splatting$_{\text{supersampled}}$ & 28.98 & 0.895 & 0.246 & 29.32 & 0.901 & 0.167 & 29.94 & 0.915 & 0.109 & 30.56 & 0.928 & 0.086 \\
\cmidrule(lr){1-13}
Ours & 28.91 & 0.894 & \tabfirst{0.244} & 29.26 & 0.900 & 0.166 & 29.77 & 0.914 & 0.110 & 30.04 & 0.926 & 0.087 \\
\midrule
\multicolumn{13}{l}{\emph{NeRF Synthetic}} \\
3DGS~\cite{kerbl3Dgaussians} & 33.46 & \tabfirst{0.970} & 0.030 & 27.00 & 0.951 & 0.032 & 21.49 & 0.880 & 0.064 & 17.84 & 0.774 & 0.117 \\
Mip-Splatting~\cite{Yu2024MipSplatting} & 33.44 & 0.969 & 0.031 & 34.01 & 0.976 & 0.019 & 31.79 & 0.977 & 0.019 & 28.58 & 0.972 & 0.026 \\
Analytic-Splatting~\cite{10.1007/978-3-031-72643-9_17} & 33.71 & \tabfirst{0.970} & \tabfirst{0.029} & 34.14 & 0.977 & 0.019 & 31.43 & 0.977 & 0.021 & 27.97 & 0.969 & 0.031 \\
Gaussian Blending~\cite{koo2026gb} & \tabfirst{33.78} & \tabfirst{0.970} & 0.030 & \tabfirst{35.58} & \tabfirst{0.979} & \tabfirst{0.017} & \tabfirst{36.16} & \tabfirst{0.984} & \tabfirst{0.013} & \tabfirst{34.77} & \tabfirst{0.985} & \tabfirst{0.014} \\
\cmidrule(lr){1-13}
Triangle Splatting$_{\text{original}}$ & 32.96 & 0.967 & 0.033 & 31.09 & 0.966 & 0.028 & 23.12 & 0.902 & 0.081 & 13.68 & 0.687 & 0.320 \\
Triangle Splatting$_{\text{no culling}}$ & 33.39 & 0.969 & 0.031 & 33.40 & 0.973 & 0.023 & 29.97 & 0.964 & 0.040 & 26.34 & 0.944 & 0.074 \\
Triangle Splatting$_{\text{supersampled}}$ & 33.40 & 0.969 & 0.032 & 35.38 & 0.978 & 0.018 & 35.78 & 0.982 & 0.015 & 33.17 & 0.980 & 0.021 \\
\cmidrule(lr){1-13}
Ours & 33.55 & 0.969 & 0.031 & 34.64 & 0.977 & 0.018 & 33.29 & 0.980 & 0.018 & 30.40 & 0.977 & 0.023 \\
\bottomrule
\end{tabular}}
\end{table*}

\begin{table*}[t]
\centering\small\setlength{\tabcolsep}{3pt}
\caption{\textbf{Zoom-out after multi-scale training} on Mip-NeRF 360~\cite{barron2022mipnerf360}, Tanks and Temples~\cite{Knapitsch2017} and Deep Blending~\cite{deep_blending}, with the scenes and resolutions of \cref{tab:comparison_ss}, and on NeRF Synthetic~\cite{mildenhall2020nerf} (8 scenes) at the four scales of its multi-scale version~\cite{barron2021mipnerf}, whose lower resolutions are box-downsampled renders. Training follows \cref{tab:comparison_ms}. Triangle Splatting$_{\text{original}}$ point-samples and culls sub-pixel triangles, and we train it at all scales with our code, since the released code has no multi-scale training. Triangle Splatting$_{\text{no culling}}$ keeps them in training and rendering, and is missing on NeRF Synthetic, since its training ran out of memory on lego. Triangle Splatting$_{\text{supersampled}}$ is trained and rendered with $2{\times}2$ samples per pixel. Analytic-Splatting and Gaussian Blending are trained with the authors' multi-scale code. The released Mip-Splatting code only trains at several scales on NeRF Synthetic, so on the other data sets we load every view at all four scales, as Analytic-Splatting does. All methods are evaluated using the same protocol, and we report the mean across scenes for each data set. \colorbox{tabfirstc}{Blue} marks the best value.}
\label{tab:supp:full_ms}
\resizebox{\textwidth}{!}{%
\begin{tabular}{lcccccccccccc}
\toprule
Method & \multicolumn{3}{c}{$1$} & \multicolumn{3}{c}{$\nicefrac{1}{2}$} & \multicolumn{3}{c}{$\nicefrac{1}{4}$} & \multicolumn{3}{c}{$\nicefrac{1}{8}$} \\
\cmidrule(lr){2-4}\cmidrule(lr){5-7}\cmidrule(lr){8-10}\cmidrule(lr){11-13}
 & PSNR$^{\uparrow}$ & SSIM$^{\uparrow}$ & LPIPS$^{\downarrow}$ & PSNR$^{\uparrow}$ & SSIM$^{\uparrow}$ & LPIPS$^{\downarrow}$ & PSNR$^{\uparrow}$ & SSIM$^{\uparrow}$ & LPIPS$^{\downarrow}$ & PSNR$^{\uparrow}$ & SSIM$^{\uparrow}$ & LPIPS$^{\downarrow}$ \\
\midrule
\multicolumn{13}{l}{\emph{Mip-NeRF 360}} \\
Mip-Splatting~\cite{Yu2024MipSplatting} & \tabfirst{27.52} & \tabfirst{0.819} & \tabfirst{0.208} & \tabfirst{28.82} & \tabfirst{0.876} & \tabfirst{0.117} & 30.36 & \tabfirst{0.922} & \tabfirst{0.065} & 31.31 & \tabfirst{0.948} & \tabfirst{0.042} \\
Analytic-Splatting~\cite{10.1007/978-3-031-72643-9_17} & 27.49 & 0.810 & 0.227 & \tabfirst{28.82} & 0.868 & 0.136 & \tabfirst{30.37} & 0.919 & 0.076 & \tabfirst{31.35} & \tabfirst{0.948} & 0.046 \\
Gaussian Blending~\cite{koo2026gb} & 26.95 & 0.800 & 0.243 & 28.17 & 0.858 & 0.153 & 29.75 & 0.912 & 0.088 & 31.11 & 0.945 & 0.052 \\
\cmidrule(lr){1-13}
Triangle Splatting$_{\text{original}}$ & 26.01 & 0.778 & 0.247 & 26.91 & 0.828 & 0.160 & 27.65 & 0.857 & 0.116 & 23.28 & 0.729 & 0.219 \\
Triangle Splatting$_{\text{no culling}}$ & 25.69 & 0.735 & 0.293 & 26.87 & 0.815 & 0.170 & 27.82 & 0.864 & 0.106 & 26.45 & 0.835 & 0.130 \\
Triangle Splatting$_{\text{supersampled}}$ & 26.65 & 0.784 & 0.236 & 27.98 & 0.854 & 0.133 & 29.56 & 0.905 & 0.074 & 30.27 & 0.921 & 0.063 \\
\cmidrule(lr){1-13}
Ours & 26.91 & 0.802 & 0.218 & 28.29 & 0.865 & 0.121 & 29.93 & 0.913 & 0.068 & 31.00 & 0.941 & 0.046 \\
\midrule
\multicolumn{13}{l}{\emph{Tanks and Temples}} \\
Mip-Splatting~\cite{Yu2024MipSplatting} & \tabfirst{23.71} & 0.843 & 0.182 & \tabfirst{24.68} & 0.893 & 0.101 & 25.66 & \tabfirst{0.930} & \tabfirst{0.056} & 26.22 & 0.947 & 0.041 \\
Analytic-Splatting~\cite{10.1007/978-3-031-72643-9_17} & 23.64 & 0.839 & 0.193 & 24.65 & 0.890 & 0.110 & \tabfirst{25.69} & 0.929 & 0.058 & \tabfirst{26.33} & \tabfirst{0.949} & \tabfirst{0.039} \\
Gaussian Blending~\cite{koo2026gb} & 23.28 & 0.836 & 0.203 & 24.19 & 0.885 & 0.120 & 25.24 & 0.925 & 0.064 & 26.08 & 0.947 & 0.041 \\
\cmidrule(lr){1-13}
Triangle Splatting$_{\text{original}}$ & 20.60 & 0.785 & 0.252 & 21.14 & 0.819 & 0.186 & 21.27 & 0.815 & 0.183 & 15.30 & 0.598 & 0.371 \\
Triangle Splatting$_{\text{no culling}}$ & 22.21 & 0.802 & 0.238 & 23.05 & 0.864 & 0.137 & 23.22 & 0.885 & 0.105 & 21.90 & 0.861 & 0.126 \\
Triangle Splatting$_{\text{supersampled}}$ & 22.85 & 0.835 & 0.188 & 23.77 & 0.889 & 0.102 & 24.68 & 0.923 & 0.064 & 24.77 & 0.931 & 0.062 \\
\cmidrule(lr){1-13}
Ours & 22.99 & \tabfirst{0.847} & \tabfirst{0.168} & 23.91 & \tabfirst{0.896} & \tabfirst{0.093} & 24.85 & 0.929 & 0.058 & 25.43 & 0.946 & 0.044 \\
\midrule
\multicolumn{13}{l}{\emph{Deep Blending}} \\
Mip-Splatting~\cite{Yu2024MipSplatting} & 29.13 & 0.900 & \tabfirst{0.250} & 29.51 & 0.913 & \tabfirst{0.155} & 30.11 & 0.930 & 0.079 & 30.67 & 0.941 & 0.056 \\
Analytic-Splatting~\cite{10.1007/978-3-031-72643-9_17} & 29.58 & 0.905 & 0.251 & 29.95 & 0.917 & \tabfirst{0.155} & 30.55 & 0.934 & \tabfirst{0.076} & 31.11 & 0.946 & \tabfirst{0.050} \\
Gaussian Blending~\cite{koo2026gb} & \tabfirst{29.59} & \tabfirst{0.907} & 0.255 & \tabfirst{29.97} & \tabfirst{0.919} & 0.161 & \tabfirst{30.63} & \tabfirst{0.936} & 0.080 & \tabfirst{31.44} & \tabfirst{0.950} & 0.052 \\
\cmidrule(lr){1-13}
Triangle Splatting$_{\text{original}}$ & 27.81 & 0.883 & 0.282 & 28.09 & 0.887 & 0.204 & 28.37 & 0.894 & 0.152 & 26.46 & 0.843 & 0.211 \\
Triangle Splatting$_{\text{no culling}}$ & 28.08 & 0.879 & 0.295 & 28.40 & 0.888 & 0.206 & 28.85 & 0.902 & 0.136 & 28.40 & 0.902 & 0.124 \\
Triangle Splatting$_{\text{supersampled}}$ & 28.62 & 0.889 & 0.265 & 29.00 & 0.899 & 0.179 & 29.66 & 0.915 & 0.110 & 30.36 & 0.929 & 0.084 \\
\cmidrule(lr){1-13}
Ours & 28.57 & 0.889 & 0.264 & 28.99 & 0.900 & 0.175 & 29.68 & 0.917 & 0.108 & 30.36 & 0.931 & 0.080 \\
\midrule
\multicolumn{13}{l}{\emph{NeRF Synthetic}} \\
Mip-Splatting~\cite{Yu2024MipSplatting} & 32.91 & 0.967 & 0.033 & 34.56 & 0.977 & 0.019 & 35.47 & 0.983 & 0.013 & 35.52 & 0.988 & 0.011 \\
Analytic-Splatting~\cite{10.1007/978-3-031-72643-9_17} & 33.21 & \tabfirst{0.968} & \tabfirst{0.032} & 34.87 & 0.977 & \tabfirst{0.018} & 35.88 & \tabfirst{0.984} & 0.013 & 35.94 & \tabfirst{0.989} & \tabfirst{0.010} \\
Gaussian Blending~\cite{koo2026gb} & \tabfirst{33.28} & \tabfirst{0.968} & 0.033 & \tabfirst{35.21} & \tabfirst{0.978} & \tabfirst{0.018} & \tabfirst{36.68} & \tabfirst{0.984} & \tabfirst{0.012} & \tabfirst{36.49} & 0.988 & \tabfirst{0.010} \\
\cmidrule(lr){1-13}
Triangle Splatting$_{\text{original}}$ & 28.41 & 0.940 & 0.068 & 29.62 & 0.950 & 0.048 & 30.32 & 0.957 & 0.040 & 25.45 & 0.930 & 0.083 \\
Triangle Splatting$_{\text{supersampled}}$ & 30.39 & 0.953 & 0.055 & 32.42 & 0.968 & 0.029 & 34.23 & 0.976 & 0.020 & 33.08 & 0.978 & 0.021 \\
\cmidrule(lr){1-13}
Ours & 31.52 & 0.960 & 0.046 & 33.61 & 0.974 & 0.023 & 35.41 & 0.982 & 0.014 & 34.98 & 0.986 & 0.012 \\
\bottomrule
\end{tabular}}
\end{table*}

\end{document}